\pdfoutput=1

\documentclass[11pt]{article}

\usepackage{coling}

\usepackage{times}
\usepackage{latexsym}

\usepackage[T1]{fontenc}

\usepackage[utf8]{inputenc}

\usepackage{microtype}

\usepackage{inconsolata}
\usepackage{amsmath}
\usepackage{graphicx}
\usepackage{hyperref}
\usepackage{url}
\usepackage{tabu}                     %
\usepackage{multirow}                 %
\usepackage{multicol}                 %
\usepackage{multirow}                %
\usepackage{float}                    %
\usepackage{makecell}                 %
\usepackage{booktabs}
\usepackage{amssymb} 
\usepackage{CJKutf8}
\usepackage{enumitem}
\usepackage{caption}
\setlist[itemize]{topsep=5pt, partopsep=1pt, parsep=1pt, itemsep=5pt} %

\title{Beyond Boundaries: Learning a Universal Entity Taxonomy across Datasets and Languages for Open Named Entity Recognition}

\author{
    \bf{\normalsize
    Yuming Yang$^{1}$,\ \
    Wantong Zhao$^{1}$,\ \
    Caishuang Huang$^{1}$,\ \
    Junjie Ye$^{1}$,\ \
    Xiao Wang$^{1}$,}\\
    \bf{\normalsize
    Huiyuan Zheng$^{1}$,\ \ 
    Yang Nan$^{1}$,\ \
    Yuran Wang$^{2}$,\ \
    Xueying Xu$^{2}$,\ \
    Kaixin Huang$^{2}$,}\\
    \bf{\normalsize
    Yunke Zhang$^{2}$,\ \
    Tao Gui$^{3,4}$\thanks{Corresponding Authors.},\ \
    Qi Zhang$^{1,5,6*}$,\ \
    Xuanjing Huang$^{1,6*}$}\\
  {$^1$ \normalsize School of Computer Science, Fudan University\ \ $^2$ \normalsize Honor Device Co., Ltd} \\
  {$^3$ \normalsize Institute of Modern Languages and Linguistics, Fudan University\ \ $^4$ \normalsize Pengcheng Laboratory} \\
  {$^5$ \normalsize Research Institute of Intelligent Complex Systems, Fudan University} \\
  {$^6$ \normalsize Shanghai Key Laboratory of Intelligent Information Processing} \\
  \texttt{\normalsize yumingyang23@m.fudan.edu.cn} \ \ \ \texttt{\normalsize \{qz,tgui,xjhuang\}@fudan.edu.cn} \\
}

\begin{document}
\maketitle
\begin{abstract}

Open Named Entity Recognition (NER), which involves identifying arbitrary types of entities from arbitrary domains, remains challenging for Large Language Models (LLMs).
Recent studies suggest that fine-tuning LLMs on extensive NER data can boost their performance.
However, training directly on existing datasets neglects their inconsistent entity definitions and redundant data, limiting LLMs to dataset-specific learning and hindering out-of-domain adaptation.
To address this, we present \texttt{B\textsuperscript{2}NERD}, a compact dataset designed to guide LLMs' generalization in Open NER under a universal entity taxonomy. 
\texttt{B\textsuperscript{2}NERD} is refined from 54 existing English and Chinese datasets using a two-step process.
First, we detect inconsistent entity definitions across datasets and clarify them by distinguishable label names to construct a universal taxonomy of 400+ entity types.
Second, we address redundancy using a data pruning strategy that selects fewer samples with greater category and semantic diversity.
Comprehensive evaluation shows that \texttt{B\textsuperscript{2}NERD} significantly enhances LLMs' Open NER capabilities.
Our B\textsuperscript{2}NER models, trained on \texttt{B\textsuperscript{2}NERD}, outperform GPT-4 by 6.8-12.0 F1 points and surpass previous methods in 3 out-of-domain benchmarks across 15 datasets and 6 languages. The data, models, and code are publicly available at \url{https://github.com/UmeanNever/B2NER}.

\end{abstract}

\section{Introduction}
\begin{figure}[t!]
    \centering
    \includegraphics[width=0.83\columnwidth]{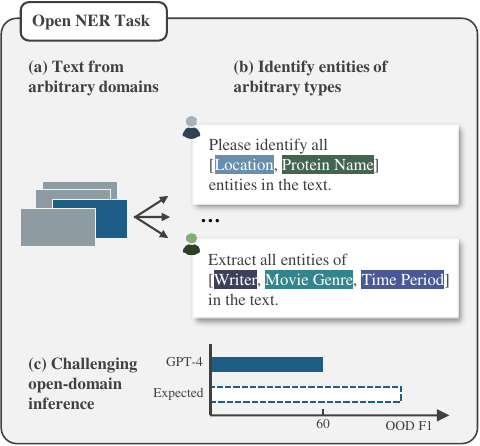}
    \caption{The Open NER task aims to extract arbitrary entities (common and unseen) from arbitrary domains (in-domain and out-of-domain). Current LLMs, like GPT-4, still fall short on this task.}
    \label{fig:openner}
\end{figure}

Open Named Entity Recognition (NER), which targets both in-domain and out-of-domain identification of common and unseen entities, is crucial for broader NER applications in real-world scenarios, such as the low-resource fields of law and biomedicine (\citealp{etzioni2008open}; \citealp{leitner2019fine}; \citealp{perera2020named}).
As shown in Figure \ref{fig:openner}, despite advancements in Large Language Models (LLMs) raising expectations for solving Open NER, current LLMs still struggle with intricate entity taxonomies in open domains and show limited NER capabilities  (\citealp{katz2023neretrieve}; \citealp{gao2023benchmarking}; \citealp{wei2023zero}; \citealp{li2023evaluating}; \citealp{Ye2023ACC}).
Recent studies (\citealp{wang2023instructuie}; \citealp{sainz2023gollie}; \citealp{xiao2023yayi}; \citealp{gui2024iepile}) address this by fine-tuning LLMs on numerous existing NER datasets, helping them learn detailed entity definitions and achieve better overall performance.

\begin{figure}[t!]
    \centering
    \includegraphics[width=1.0\columnwidth]{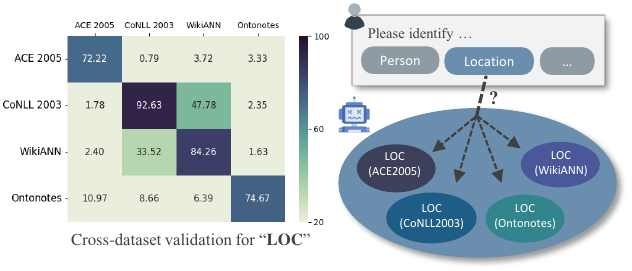}
    \caption{Sample results of BERT-based cross-dataset entity validation for \textsc{LOC} entity. Light colors indicate conflict entity definitions. Training LLM on these inconsistent datasets leads to confusions during inference.}
    \label{fig:conflicts}
\end{figure}

However, directly using existing datasets to train Open NER models is hindered by two flaws that limit the models' out-of-domain generalization:
(1) \textbf{Inconsistent and vague entity definitions across datasets.}
Different datasets often have conflicting entity definitions. 
For instance, some datasets distinguish between locations like "Times Square" and geopolitical entities like "Paris", while others annotate both as \textsc{LOC}.
Aligning LLMs with these inconsistencies leads to dataset-specific patterns and confusion over common entities during inference (Figure \ref{fig:conflicts}).
To avoid conflicts, \citealp{zhou2023universalner} suggests adding dataset names in training prompts, but this cannot improve out-of-domain inference with unknown datasets.
\citealp{sainz2023gollie} introduces detailed annotation guidelines for each entity type, but such guidelines are hard to obtain and challenging for LLMs to understand.
(2) \textbf{Redundant data in combined datasets.}
Most datasets heavily annotate common entities, with fewer samples for long-tail entities. Thus, the combined dataset contains redundant samples with similar annotations and semantics.
This lack of diversity may cause LLMs to overfit and hinder universal generalization (\citealp{zhou2023lima}; \citealp{liu2024what}).
To circumvent above issues, some studies (\citealp{zhou2023universalner}; \citealp{li2024knowcoder}; \citealp{ding2024rethinking}) explore using synthetic NER data annotated by ChatGPT, but synthetic data struggles to meet real-world NER requirements. 
The valuable human annotations in existing datasets remain underutilized.

In this work, we propose enhancing LLMs for Open NER by directly addressing issues in existing training datasets and normalizing them into a compact collection via a two-step approach.
First, we systematically standardize entity definitions across all collected datasets. 
Inconsistent entity definitions are automatically detected via model-based validation and rule-based screening. 
We then clarify these ambiguous entity types by assigning distinguishable label names for each unique type following detailed principles.  
This step forms our universal entity taxonomy, which guides the categorization of both common and unseen entities.
Second, we avoid redundancy by employing a data pruning strategy that considers both category and semantic diversity. 
Our strategy samples equally from each entity type and selects samples with lower textual similarity within each type to enhance semantic diversity. 
By applying above approach on our bilingual (English and Chinese) NER collection of 54 datasets, we derive \texttt{B\textsuperscript{2}NERD}, a \textbf{B}eyond-\textbf{B}oundary \textbf{NER} \textbf{D}ataset with a universal taxonomy of 400+ entity types across 16 major domains.

By fine-tuning on \texttt{B\textsuperscript{2}NERD}, we develop B\textsuperscript{2}NER models --- LLMs with extensive Open NER capabilities that generalize across datasets and languages.
Experimental results on 3 out-of-domain NER benchmarks across 15 datasets show that our model outperforms both GPT-4 and previous methods by 3.0\% in English, 6.8\% in Chinese, and 6.7\% in a multilingual setting. Further analysis offers deeper insights into our approach's effectiveness.

Our main contributions are three-fold:
\begin{itemize}
\item We present \texttt{B\textsuperscript{2}NERD}, a cohesive and compact dataset that advances LLM capabilities for Open NER, along with its full version, \texttt{B\textsuperscript{2}NERD\textsubscript{all}}, the largest bilingual NER data collection to date.
\item We introduce a two-step approach to address the inconsistencies and redundancy among existing NER datasets, creating a universal entity taxonomy that transcends dataset boundaries.
\item Experiments show that our B\textsuperscript{2}NER models outperform GPT-4 and previous methods in comprehensive out-of-domain evaluations across various datasets and languages.
\end{itemize}

\renewcommand{\UrlFont}{\footnotesize}

\begin{figure*}
    \centering
    \includegraphics[width=1.94\columnwidth]{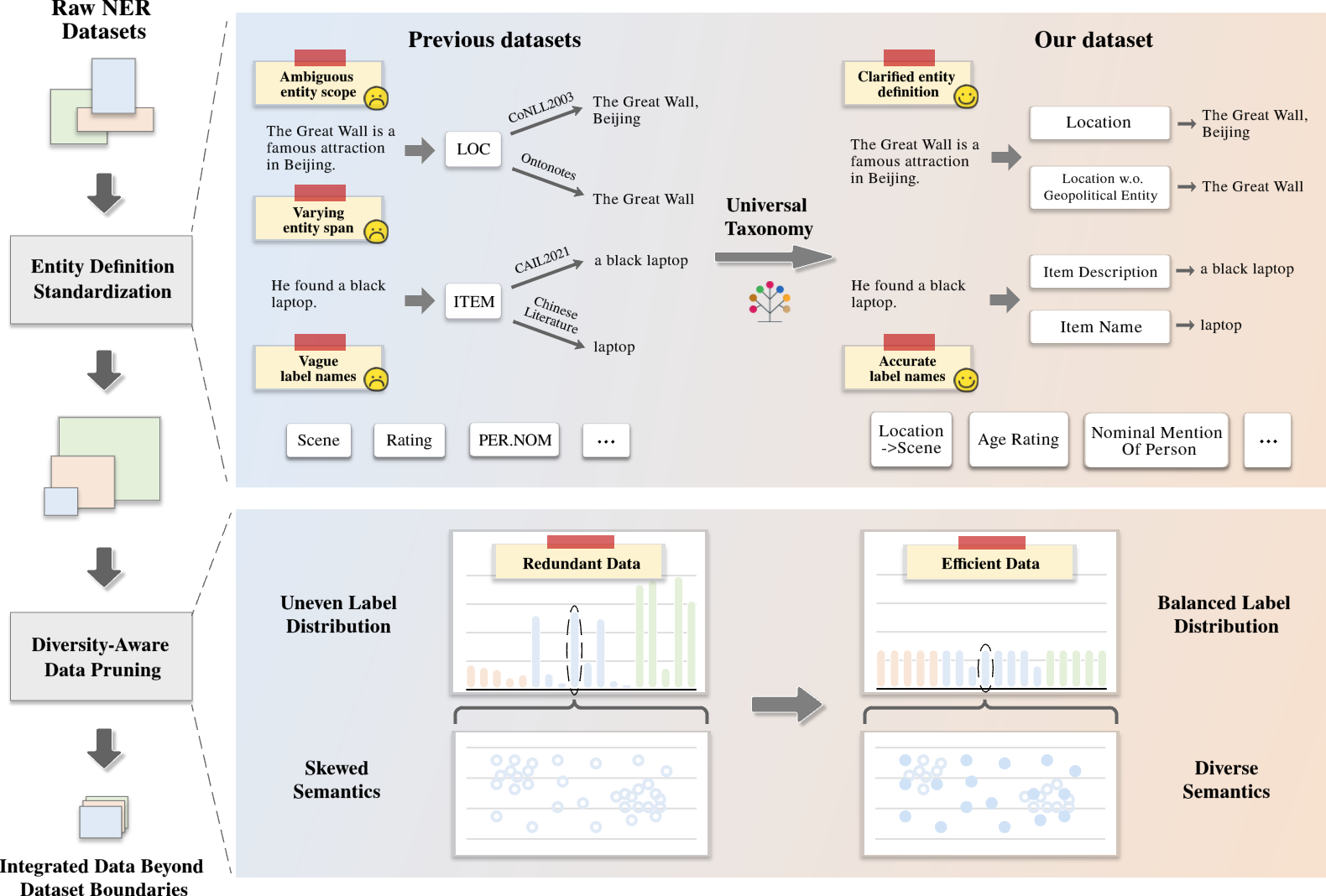}
    \caption{Framework of \texttt{B\textsuperscript{2}NERD} data construction: raw NER datasets are reshaped into a cohesive dataset via entity definition standardization and diversity-aware data pruning. Final data is then used to train our Open NER model.}
    \label{fig:method}
\end{figure*}

\section{Preliminaries}
We first discuss our data collection process and the limitations of using collected datasets.

\subsection{Data Collection}
\label{sec:dcc}

To meet the diverse needs of Open NER, we gather the largest collection of existing datasets.
For English NER, we use the collection from \citealp{wang2023instructuie}. 
For Chinese NER, we invest extensive effort in data collection due to the limited datasets in prior work. More details are in Appendix \ref{app:dcc}. 
Finally, we derive a bilingual collection of 54 datasets from 16 major domains, as shown in Figure \ref{fig:dataset}. 

\subsection{Inconsistencies among Collected Datasets}
\label{sec:icd}
The collected datasets have differing entity definitions.
To quantify their conflicts, we conduct cross-dataset entity validation experiments.
Figure \ref{fig:conflicts} shows a sample experiment among 4 datasets, all having \textsc{LOC} entities.
We iteratively train a BERT-based model on one dataset and evaluate its performance on recognizing \textsc{LOC} entities in the other three.
Low F1 scores indicate inconsistent entity definitions between datasets. 
The results reveal significant conflicts among collected datasets, which confuse LLMs during training and inference.
More explanations are provided in Appendix \ref{app:autod}.

\section{Approach}

We propose a two-step approach to address existing dataset inconsistencies and redundancy. This section details the construction of the \texttt{B\textsuperscript{2}NERD} dataset (Sections \ref{EDS}-\ref{DDS}) and the training of B\textsuperscript{2}NER models (Section \ref{IT}). Figure \ref{fig:method} outlines our framework.

\subsection{Entity Definition Standardization with Universal Taxonomy}
\label{EDS}
As shown in Figures \ref{fig:conflicts} and \ref{fig:method}, the same entity label often has different meanings across datasets, and many labels are unclear outside their original context.
To address these ambiguities and avoid dataset-specific learning, we systematically standardize entity definitions in existing datasets by detecting conflicts, clarifying ambiguous entities for a universal taxonomy, and renaming entity labels.
New entities can also be easily accommodated within this taxonomy following our practice.

\begin{itemize}[label={}, labelsep=0pt, leftmargin=0pt]  
    \item \textbf{Automatic dataset conflict detection.} \hspace{0.15cm}
    First, we detect conflicts among datasets at scale by identifying inconsistent annotations for entities with similar label names using automatic methods:
    \textbf{Model-based cross validation}: We extend the method in Section \ref{sec:icd} to all dataset pairs with similar entity types, identifying potential conflict entity definitions from low F1 results.
    \textbf{Rule-based screening}: 
    To further understand these conflicts, we screen for cases when same entity mention receives different annotations across datasets.
    Significant inconsistent cases are classified and listed for future processing.
    See Appendix \ref{app:autod} for more details.

    \item \textbf{Resolving Conflicts and Constructing Universal Taxonomy.} \hspace{0.15cm}
    For entities with similar label names but different definitions, we invite experts to scrutinize their differences and split them into unique entity types, following NER guidelines like ACE\footnote{\url{https://www.ldc.upenn.edu/sites/www.ldc.upenn.edu/files/english-entities-guidelines-v6.6.pdf}}.
    As partially shown in Figure \ref{fig:method}, we address major issues like:
    (1) \textbf{Different entity scope}: The same label name might encompass a different range of entities, as shown in the \textsc{LOC} example in Figure \ref{fig:method}.
    (2) \textbf{Different entity span}: Different datasets may identify different spans for the same entity label, as shown in the \textsc{Item} example in Figure \ref{fig:method}.
    (3) \textbf{Different mention type}: 
    There are various ways to refer an entity.
    For \textsc{PER} entities, most datasets recognize explicit names like "Jack Smith", while others, like \texttt{ACE 2005}, include nominal or pronoun mentions like "a hunter" and "he" as \textsc{PER}. 
    The latter will be split as a new entity type \textsc{General Mentions of Person}.
    (4) \textbf{Overlapped entity granularity}:
    When a dataset contains both coarse and fine-grained entities, like \textsc{Person} and \textsc{Writer}, the model may only label the coarse type (\citealp{sainz2023gollie}). 
    We believe the former type actually refers to "other person" in such datasets and should be distinguished as \textsc{Person->Others}.
    By applying this clarification and separation process to each entity type, we create a universal entity taxonomy with consistent definitions across all datasets.

\begin{figure}[t!]
    \centering
    \includegraphics[width=0.92\columnwidth]{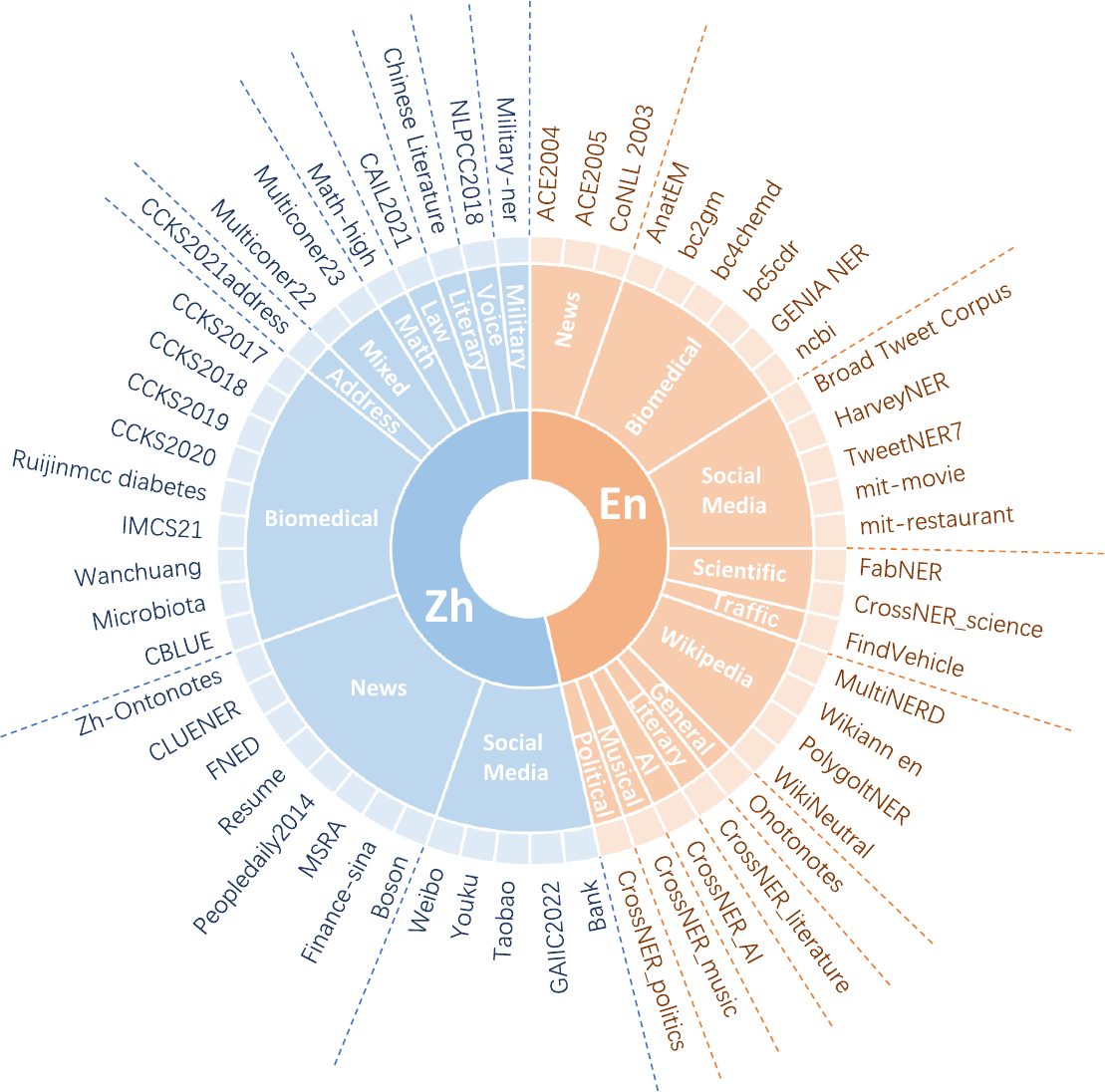}
    \caption{Overview of collected datasets in \texttt{B\textsuperscript{2}NERD}. We collected 54 datasets across 16 major domains.}
    \label{fig:dataset}
\end{figure}

    \item \textbf{Reassigning Label Names.} \hspace{0.15cm}
    Label names are crucial in generative NER as they appear in both prompts and answers, depicting current task.
    Thus, we reassign natural language labels to entity types in the universal taxonomy to more accurately represent their definitions for LLM understanding.
    Our naming system adheres to the following principles:
    \textbf{Readable}: Labels should be clear words or phrases, avoiding acronyms.
    \textbf{Unambiguous}: Each label should distinctly differentiate between similar entity types, like \textsc{Item name} v.s. \textsc{Item description}.
    \textbf{Hierarchical}: Entity sub-types are named after their parent types (e.g., \textsc{Location->Scene}), aiding in polysemous resolution and clear granularity levels.
    \textbf{Flexible}: To accommodate diverse NER tasks and new entities types, the system allows adaptable naming, such as using "or" in labels.
    For example, a type encompassing both person and group entities is labeled \textsc{Person or Group}.
\end{itemize}

The clarifying and renaming process is performed by human experts to ensure higher data quality. 
We publicly recruit college students with sufficient NER annotation experience and implement procedures such as preliminary training and attention checks to maintain overall work quality and consistency (see Appendix \ref{app:EDS} for details). 
The final taxonomy includes 400+ diverse entity types, as shown in Table \ref{tbl:taxonomy}. 
The standardized datasets with consistent entities comprise our \texttt{B\textsuperscript{2}NERD\textsubscript{all}} collection, containing the full data.

\begin{table*}[t!]
    \centering
    \resizebox{1.99\columnwidth}{!}{
    \begin{tabular}{c|c|c|c|c}
        \Xhline{1.5pt}
        \textbf{Person} & \textbf{Location} & \textbf{Organization} & \textbf{Object} & \textbf{Creative Work}\\

        \hline
\rule{0pt}{2.6ex} 
Person&Location&Organization&Astronomical Object&Creative Work\\

Person->Artist&Location->Facility&Organization->Sport team&Product Name&Creative Work->Album\\

Person->Scientist&Location->Starting point&Organization->Corporation&Product Name->Music Instrument&Creative Work>Literature\\

Person->Victim&Geo-Political entity&Organization->Political Group&Product Name->SUV&Creative Work->Magazine\\

General Mention of Person&Location(w/o Country)&Organization->Band&Crime Tool&Creative Work->Software\\

Nationality&City&Organization->Public Company&Item Description&Music Theme\\

...&...&...&...&...\\
        \Xhline{1.5pt}
        \textbf{Measure} & \textbf{Time}& \textbf{Education} & \textbf{Biomedical} & \textbf{Other}\\
        \hline
\rule{0pt}{2.6ex} 
Cardinal Number&Date or Period&Academic Major&Anatomy&Event Name\\

Dose&Sub-day Time Expression&Research Field&Drug or Vaccine&Mobile Phone Model\\

Measurement Quantity&Duration&Education Background&Chemical&Legal Document\\

Monetary Amount(with unit)&Generation&Mathematical concept&Microorganism&Financial Term\\

Property Value&Operating Hours &Mathematical principle or method&Disease Name&Type Of Emotion\\

Percentage(with \%)&Frequency&Academic Conference&Symptom&Color\\

...&...&...&...&...\\
        \Xhline{1pt}
    \end{tabular} }
    \caption{The universal entity taxonomy for \texttt{B\textsuperscript{2}NERD} includes 400+ entity types across 10+ main categories. New entities can be easily accommodated within this taxonomy. The full table is available in Appendix \ref{app:dst}.}
    \label{tbl:taxonomy}
\end{table*}

\subsection{Diversity-aware Data Pruning}
\label{DDS}

Despite addressing inconsistencies, the merged dataset retains imbalanced data distribution from raw datasets. 
For instance, \textsc{Location} entities in news are heavily annotated, while long-tail entities like \textsc{City} are sparse.
To avoid model over-fitting to redundant data, we propose fine-tuning LLMs on a curated, diverse subset of \texttt{B\textsuperscript{2}NERD\textsubscript{all}} to learn more transferable patterns.

As depicted in the lower part of Figure \ref{fig:method}, we address diversity in NER datasets by ensuring \textbf{balanced data distribution} across entity types and \textbf{diverse text semantics} within each type's samples.

To maximize diversity in a limited-size dataset, our strategy selects $k$ semantically diverse samples per unique entity type. 
We start by initializing sample pools for all entity types $S_{1},\cdots,S_{M} = \emptyset$, each holding up to $k$ samples.
Then for each random sample $x$ with annotations of entity type $(t_1, t_2, ... t_M)$, we check the status of related pools ($S_{t_1}, S_{t_2}, ..., S_{t_M}$). 
For non-full pools, we assess the maximum semantic similarity between $x$ and corresponding sample pool $S_{t_i}$. 
We decide whether to add $x$ to this pool based on the probability: 
\begin{align*}
p(S_{t_i} \leftarrow S_{t_i} &\cup \{x\}) = 1 - \max_{y \in S_{t_i}} \text{sim}(x, y) \\
&= 1 - \max_{y \in S_{t_i}} \text{cosine}(x, y) + b
\end{align*}
The offset $b$ controls the penalty for semantic similarity.
We utilize AnglE (\citealp{li2023angle}) to generate text embeddings for samples and calculate semantic distance using cosine similarity between their embeddings.
This approach prioritizes semantically diverse samples.
Notably, if a sample is added to a pool, all entity mentions within it are retained for optimal data efficiency. 
The sampling process continues until all pools are full or all samples have been traversed. 
The final selected dataset is formed by combining unique samples from all pools.
In practice, we treat the same entity type in different datasets as distinct entity types, enabling efficient per-dataset pruning.
Additionally, up to $\frac{1}{5}k$ negative samples are added per dataset.

We implement the diversity-aware data pruning strategy on the training set of \texttt{B\textsuperscript{2}NERD\textsubscript{all}} using $k=400$ and $b=0$.
In Section \ref{subsec:analysis}, we compare different sampling methods and data scales.
The resulting \texttt{B\textsuperscript{2}NERD} dataset (Table \ref{tbl:dataset_statistics}), featuring a universal NER taxonomy and efficient samples, enables LLMs to generalize beyond raw dataset boundaries.

\begin{table}[t!]
\centering
\resizebox{0.88\columnwidth}{!}{
\begin{tabular}{llllll}
\toprule
\textbf{Split} & \textbf{Lang.} & \textbf{Datasets} & \textbf{Types} & \textbf{Num} & \textbf{Raw Num} \\
\midrule
\multirow{3}{*}{Train} & En & 19 & 119 & 25,403 & 838,648  \\
 & Zh & 21 & 222 & 26,504 & 580,513 \\ \cmidrule{2-6}
 & Total & 40 & 341 & 51,907 & 1,419,161 \\ 
 \midrule
\multirow{3}{*}{Test} & En & 7 & 85 & - & 6,466 \\
 & Zh & 7 & 60 & - & 14,257\\ \cmidrule{2-6}
 & Total & 14 & 145 & - & 20,723\\
\bottomrule
\end{tabular}
}
\caption{Dataset statistics for \texttt{B\textsuperscript{2}NERD}. "Num" refers to the number of samples in the final \texttt{B\textsuperscript{2}NERD}. "Raw Num" represents the samples in collected datasets and \texttt{B\textsuperscript{2}NERD\textsubscript{all}} before data pruning.}
\label{tbl:dataset_statistics}
\end{table}

\subsection{Instruction Tuning with Regularization}
\label{IT}

Based on \texttt{B\textsuperscript{2}NERD}, we conduct instruction tuning on LLMs to create the B\textsuperscript{2}NER models, using a UIE-style NER instruction template similar to \citealp{wang2023instructuie} (See Appendix \ref{app:it}). 

We observe that instructions for samples from the same dataset include a shared part about entity label options ("Label Set:[...]"), causing LLMs to mechanically memorize this part rather than understanding the actual labels.
Addressing this, we introduce training regularization methods to prevent dataset-specific patterns in the instructions.
A key innovation is \textbf{Dynamic Label Set}: Instead of asking LLMs to recognize a fixed set of labels for each dataset, we randomly vary the number and order of entity types mentioned in the prompts to reduce co-occurrences. See Appendix \ref{app:trainreg} for more training regularization details.

\begin{table*}[t!]
\centering
\resizebox{1.99\columnwidth}{!}{
\begin{tabular}{l|cc|ccccc|c|c}
\toprule
 & \multicolumn{2}{c|}{\emph{w/ Unseen Entities}} & \multicolumn{5}{c|}{\emph{w/ Common \& Unseen Entities}} & &  \\
\textbf{Model} & \textbf{Movie} & \textbf{Restaurant} & \textbf{AI} & \textbf{Litera.} & \textbf{Music} & \textbf{Politics} & \textbf{Science} & \textbf{Avg.} & \textbf{Instance/s}\\ 
\midrule
\emph{Non-Natural Language Prompt} & & & & & & & & & \\
\hspace{0.25cm} GoLLIE-7B         & 63.0 & 43.4 & 59.1 & 62.7 & 67.8 & 57.2 & 55.5 & 58.4 & - \\ 
\hspace{0.25cm} KnowCoder-7B      & 50.0 & 48.2 & 60.3 & 61.1 & 70.0 & 72.2 & 59.1 & 60.1 & - \\ 
\hspace{0.25cm} GNER-7B           & 68.6 & 47.5 & 63.1 & 68.2 & 75.7 & 69.4 & 69.9 & 66.1 & 4.0\\
\hspace{0.25cm} GNER-11B          & 62.5 & 51.0 & \textbf{68.2} & 68.7 & \underline{81.2} & 75.1 & \underline{76.7} & 69.1 & 3.0\\ 
\emph{Natural Language Prompt} & & & & & & & & & \\
\hspace{0.25cm} InstructUIE-11B   & 63.0 & 21.0 & 49.0 & 47.2 & 53.2 & 48.2 & 49.3 & 47.3 & 3.4\\ 
\hspace{0.25cm} UniNER-7B         & 59.4 & 31.2 & 62.6 & 64.0 & 66.6 & 66.3 & 69.8 & 60.0 & 1.6\\  
\hspace{0.25cm} GPT-4             & 60.4 & \textbf{59.7} & 50.0 & 55.2 & 69.2 & 63.4 & 63.2 & 60.1 & - \\ 
\midrule
\hspace{0.25cm} Baseline-7B       & 49.7 & 36.6 & 43.7 & 44.0 & 58.6 & 59.8 & 60.0 & 50.3 & - \\
\hspace{0.25cm} B\textsuperscript{2}NER-7B (w/o English) & 68.5 & 50.4 & 56.9 & 55.0 & 65.1 & 67.2 & 65.9 & 61.3 & - \\
\hspace{0.25cm} B\textsuperscript{2}NER-7B (only English) & 67.6 & 53.3 & 59.0 & 63.7 & 68.6 & 67.8 & 72.0 & 64.6 & - \\
\hspace{0.25cm} B\textsuperscript{2}NER-7B                & \underline{70.2} & 56.8 & 64.1 & \underline{69.0} & 76.4 & \underline{75.5} & \underline{76.7} & \underline{69.8} & \textbf{16.1} \\
\hspace{0.25cm} B\textsuperscript{2}NER-20B & \textbf{71.4} & \underline{57.1} & \underline{64.7} & \textbf{71.6} & \textbf{82.4} & \textbf{78.2} & \textbf{79.4} & \textbf{72.1} & \underline{7.0}\\
\bottomrule
\end{tabular} }
\caption{Out-of-domain evaluation results on English NER. \textbf{Bold} numbers highlight the best scores, while \underline{underlined} numbers indicate suboptimal scores. "\emph{w/ Unseen Entities}" denotes datasets with every entity type unseen during training. "\emph{w/ Common \& Unseen Entities}" denotes datasets with a mix of common and unseen entities. "w/o English" refers to a cross-lingual model trained without any English data. Our models use InternLM2 as the backbone LLM; results on other backbones are shown in Appendix \ref{app:backb}. The last column reports inference speed, tested on a single 8$\times$A100 node with a batch size of 4 per device following \citealp{ding2024rethinking}. }
\label{tbl:main-en}
\end{table*}

\begin{table*}[t!]
\centering
\resizebox{1.99\columnwidth}{!}{
\begin{tabular}{l|ccc|cccc|cc}
\toprule
 & \multicolumn{3}{c|}{\emph{w/ Unseen Entities}} & \multicolumn{4}{c|}{\emph{w/ Common \& Unseen Entities}} & &  \\
\multirow{2}{*}{\textbf{Model}} & \textbf{Law} & \textbf{Math} & \textbf{Address} & \textbf{Cluener} & \textbf{Medical} & \textbf{Weibo} & \textbf{Onto. 4} & \multicolumn{2}{c}{\textbf{Average}} \\ 
 & \small (CAIL2021) & & \small (CCKS2021) & & \small (CBLUE) & & & \small (SoTA) & \small (All) \\ 
\midrule
SoTA & - & - & 68.5* & 36.5\textsuperscript{$\dagger$} & 31.4* & 38.0\textsuperscript{$\ddagger$} & 39.2\textsuperscript{$\S$} & 42.7 & - \\
GPT-4 & \textbf{69.1} & 45.9 & 70.5 & 55.7 & 44.6 & 34.0 & 68.8 & 54.7 & 55.5 \\ 
\midrule
Baseline-7B & 52.0 & 44.5 & 65.5 & 55.7 & 43.3 & 33.0 & 73.8 & 54.3 & 52.5\\
B\textsuperscript{2}NER-7B (w/o Chinese)  & 58.7 & 60.2 & 56.6 & 51.7 & 43.7 & 38.6 & 70.7 & 52.3 & 54.3 \\
B\textsuperscript{2}NER-7B (only Chinese) & 66.6 & 50.9 & 68.4 & 57.1 & 46.1 & 39.5 & 76.3 & 57.5 & 57.9 \\
B\textsuperscript{2}NER-7B                & 64.7 & 60.8 & \textbf{73.0} & 60.3 & 45.0 & 41.3 & 77.4 & 59.4 & 60.3 \\
B\textsuperscript{2}NER-20B               & 67.6 & \textbf{62.2} & 71.0 & \textbf{64.4} & \textbf{46.8} & \textbf{44.6} & \textbf{79.8} & \textbf{61.3} & \textbf{62.3} \\
\bottomrule
\end{tabular}
}
\caption{Out-of-domain evaluation results on Chinese NER. "*", "$\dagger$", "$\ddagger$", and "$\S$" denote the SoTA results from \citealp{fang2023manner}, YAYI-UIE (\citealp{xiao2023yayi}), IEPile (\citealp{gui2024iepile}), and \citealp{xie2023empirical}, respectively. }
\label{tbl:main-zh}
\end{table*}

\section{Experimental Settings}

\subsection{Implementation}
\label{subsec:setup}

\begin{itemize}[label={}, labelsep=0pt, leftmargin=0pt]
    \item \textbf{Data} \hspace{0.15cm} Our B\textsuperscript{2}NER models are trained on the \texttt{B\textsuperscript{2}NERD} dataset, containing 25,403 English samples and 26,504 Chinese samples. 
    For comparison with previous works, we include \texttt{Pile-NER} from \citealp{zhou2023universalner} as extra training data. 
    Test datasets are held out for out-of-domain evaluation in Section \ref{subsec:eva}. 
    More statistics are in Appendix \ref{app:dst}.
    
    \item \textbf{Backbone} \hspace{0.15cm} We derive our models by fine-tuning InternLM2 (\citealp{cai2024internlm2}) with LoRA (\citealp{hu2021lora}). Training details are in Appendix \ref{app:train}. InternLM2 is chosen for its balanced performance in English and Chinese, fitting our bilingual training data. We also validate our approach using other backbones in Appendix \ref{app:backb}.
\end{itemize}

\subsection{Evaluation}
\label{subsec:eva}

\begin{itemize}[label={}, labelsep=0pt, leftmargin=0pt]
    \item \textbf{Benchmarks} \hspace{0.15cm} 
    As the core aspect of the Open NER task, we assess the model's out-of-domain performance on 3 benchmarks using held-out datasets from the training data. 
    For English NER (Table \ref{tbl:main-en}), we follow \citealp{wang2023instructuie} and use 7 datasets from CrossNER and MIT.
    For Chinese NER (Table \ref{tbl:main-zh}), we create a comprehensive OOD benchmark by holding out 7 Chinese datasets covering various domains and entity types.
    For multilingual NER (Table \ref{tbl:main-cross}), we use \texttt{Multiconer22} (\citealp{malmasi-etal-2022-multiconer}) to evaluate cross-lingual effects. 
    These held-out datasets include both unseen and common entities, reflecting practical scenarios. 
    Datasets with all unseen entity types, representing a stricter zero-shot evaluation, are highlighted in our result tables.
    We also conduct in-domain supervised evaluation on 20 English datasets from \citealp{wang2023instructuie} and 6 Chinese datasets from our collection. 
    
    \item \textbf{Metrics} \hspace{0.15cm} Evaluation is based on strict span-based micro-F1, requiring exact entity type and boundary matching. Experiments are repeated four times, and the results are averaged.
\end{itemize}

\subsection{Compared Systems}

For English NER, we primarily compare our model with \textbf{InstructUIE} (\citealp{wang2023instructuie}) and \textbf{UniversalNER} (\citealp{zhou2023universalner}), which use similar training data and natural language prompts to us. We also include strong generative NER systems that don't use natural language prompts, such as code-based \textbf{GoLLIE} (\citealp{sainz2023gollie}), \textbf{KnowCoder} (\citealp{li2024knowcoder}) and BIO tag-based \textbf{GNER} (\citealp{ding2024rethinking}). Additionally, we train a \textbf{Baseline} model with the same data sources and backbone as our B\textsuperscript{2}NER models but without dataset normalization approaches.

For Chinese NER, we compare with SoTA zero-shot or OOD NER systems including \citealp{xie2023empirical}, \textbf{YAYI-UIE} (\citealp{xiao2023yayi}) and \textbf{IEPile} (\citealp{gui2024iepile}). We also include 1-shot NER results from \citealp{fang2023manner}. 

Moreover, we compare our models with \textbf{GPT-4} (\citealp{achiam2023gpt}), a milestone proprietary LLM. For both English and Chinese, we prompt GPT-4-0613 to perform entity recognition on test datasets using the same instructions and standardized label names as our model. To ensure fairness, we fix format issues in GPT-4's responses.

\section{Experiment Results}

\subsection{Out-of-domain Evaluation}
Comprehensive experiments across languages and datasets demonstrate our method's effectiveness in improving out-of-domain generalization.

\begin{itemize}[label={}, labelsep=0pt, leftmargin=0pt]

\item \textbf{English NER} \hspace{0.15cm} Table \ref{tbl:main-en} shows the out-of-domain evaluation results on the English NER benchmark.
Both B\textsuperscript{2}NER-7B and B\textsuperscript{2}NER-20B exhibit superior average performance over previous methods and surpass GPT-4 by 9.7--12.0 F1 points, demonstrating their advanced capabilities.
Compared to Baseline, InstructUIE, and UniNER, which use similar data sources and prompts, B\textsuperscript{2}NER-7B significantly improves for all 7 datasets with unseen or mixed entities, highlighting the value of our normalized data.
Moreover, B\textsuperscript{2}NER-7B (69.8\%) slightly surpasses the previous SoTA GNER-11B (69.1\%) despite its smaller size and achieves a much faster inference speed (4X) than GNER-7B. 
This speed stems from our generic UIE-style prompt (See Appendix \ref{app:it}) that extracts only relevant content, unlike GNER's prompts that generate all text with tags, leading to longer responses and less flexibility. 
Additionally, we observe a surprising cross-lingual effect: our "w/o English" model trained without any English data achieves comparable performance to GPT-4 on English, showing that the learned universal taxonomy can transfer between languages.

\item \textbf{Chinese NER} \hspace{0.15cm} Table \ref{tbl:main-zh} presents the out-of-domain evaluation results on our Chinese NER benchmark. 
Both our 7B and 20B models outperform GPT-4 and other methods, exceeding the previous SoTA by 18.6 points on average.
B\textsuperscript{2}NER-7B substantially improves upon the Baseline model for all 7 datasets with unseen or mixed entities, further validating the value of our normalized data.
Moreover, B\textsuperscript{2}NER-7B boosts the average performance of the "only Chinese" model on Chinese and the "only English" model on English, showing that joint training with our bilingual NER data enhances performance in both languages. 
This suggests our universal taxonomy addresses the data disparity concerns of bilingual training, as discussed by \citealp{gui2024iepile}.

\item \textbf{Multilingual NER} \hspace{0.15cm} 
Table \ref{tbl:main-cross} shows the out-of-domain evaluation results on the multilingual dataset \texttt{Multiconer22}. 
We include 6 languages that constitute more than 0.1\% of the general LLM pretraining corpus (\citealp{touvron2023llama}). 
For strict OOD evaluations, we exclude all \texttt{Multiconer22} and \texttt{Multiconer23} samples from our training data.
We compare our model with ChatGPT (evaluated by \citealp{Lai2023ChatGPTBE}) and GLiNER (\citealp{Zaratiana2023GLiNERGM}), which uses mdeBERTa-v3-base as backbone.
From the results, our model achieves the best performance on 5 out of 6 languages. 
In the cross-lingual setting, without any training data in the target languages, our method improves the baseline model from 28.2\% to 39.9\%, outperforming other unsupervised methods and showing that learning a universal taxonomy benefits LLM generalization across language boundaries.

\end{itemize}

\begin{table}[t!]
\centering
\resizebox{0.98\columnwidth}{!}{
\begin{tabular}{lc|cc|cc}
\toprule
\textbf{Language} & \textbf{Sup.} & \textbf{ChatGPT} & \textbf{GLiNER} & \textbf{Base.} & \textbf{Ours-7B} \\ 
\midrule
\hspace{0.15cm} English & 62.7 & 37.2 & 41.7 & 39.8 & \textbf{54.8} \\ 
\hspace{0.15cm} Chinese & 53.1 & 18.8 & 24.3 & 32.8 & \textbf{45.4} \\
\midrule
\multicolumn{2}{l|}{\emph{Cross-Lingual}} & & & & \\
\hspace{0.15cm} German & 64.6 & 37.1 & \textbf{39.5} & 26.5 & 36.6\\ 
\hspace{0.15cm} Spanish & 58.7 & 34.7 & 42.1 & 34.1 & \textbf{46.0}\\ 
\hspace{0.15cm} Dutch & 62.6 & 35.7 & 38.9 & 32.2 & \textbf{43.0}\\
\hspace{0.15cm} Russian & 59.7 & 27.4 & 33.3 & 19.9 & \textbf{33.9}\\ 
\midrule
Average\textsubscript{cross} & 61.4 & 33.7 & 38.5 & 28.2 & \textbf{39.9}\\ 
Average\textsubscript{all} & 60.2 & 31.8 & 36.6 & 30.9 & \textbf{43.3} \\ 
\bottomrule

\end{tabular}
}
\caption{Out-of-domain multilingual evaluation results on \texttt{multiconer22}. "Sup." indicates supervised baseline results from \citealp{malmasi-etal-2022-multiconer}. "Base." denotes the baseline model trained without dataset normalization.}
\label{tbl:main-cross}
\end{table}

\subsection{In-domain Supervised Evaluation}
\label{subsec:id}

\begin{table}[t!]
    \centering
    \resizebox{0.88\columnwidth}{!}{
    \begin{tabular}{lcc|c}
        \toprule
\multirow{2}{*}{\textbf{Model}} & \textbf{EN Avg.} & \textbf{ZH Avg.} & \textbf{Avg.}  \\
 &  (20 Datasets) & (6 Datasets) & \\
\midrule

BERT-based                & 80.09 & 84.74 & 82.42 \\
\midrule
InstructUIE-11B           & 81.16 & - & - \\
UniNER-7B                 & \textbf{84.78} & - & - \\
\midrule
B\textsuperscript{2}NER-7B & 83.85 & \textbf{85.11} & \textbf{84.48} \\
        \bottomrule
    \end{tabular} }
    \caption{In-domain supervised evaluation results on 20 English and 6 Chinese datasets. "EN" and "ZH" denote results on English and Chinese, respectively. The full table with details can be found in Appendix \ref{app:idmore}.}
    \label{tbl:id}
\end{table}

Despite our focus on out-of-domain generalization, we also conduct in-domain supervised experiments. 
In this setting, we train and evaluate our B\textsuperscript{2}NER model on 20 English datasets (\citealp{wang2023instructuie}) and 6 Chinese datasets from our \texttt{B\textsuperscript{2}NERD\textsubscript{all}} collection with standardized entity labels.
Following previous work, we sample 10,000 examples for each dataset instead of using our pruning strategy.
The training arguments slightly differ from those used in OOD experiments (see Appendix \ref{app:train}). 
We compare our model with BERT-based task-specific models, using English results from \citealp{wang2023instructuie} and Chinese results from our evaluation. 

Results are shown in Table \ref{tbl:id}. B\textsuperscript{2}NER-7B achieves better average performance than BERT-based models on both English and Chinese datasets. For the 20 English datasets, B\textsuperscript{2}NER-7B outperforms InstructUIE-11B and slightly trails UniNER-7B by 1 point. These results demonstrate that our approach holistically enhances Open NER capabilities, achieving both superior out-of-domain generalization and competitive in-domain performance.

\subsection{Ablation Study}
\label{subsec:abl}

\begin{table}[t!]
    \centering
    \resizebox{0.92\columnwidth}{!}{
    \begin{tabular}{lcc|c}
        \toprule
        \textbf{Model} & \textbf{EN} & \textbf{ZH} & \textbf{OOD Avg.} \\ 
        \midrule

B\textsuperscript{2}NER-7B & 69.8 & 60.3 & 65.1 \\

\hspace{0.20cm} w/o entity definition std. & 62.4 & 58.5 & 60.5\textsubscript{$\downarrow$4.6} \\
\hspace{0.60cm} w/ dataset names & 60.5 & 57.0 & 58.8\textsubscript{$\downarrow$6.3} \\
\hspace{0.20cm} w/o data pruning & 66.2 & 56.6 & 61.4\textsubscript{$\downarrow$3.7} \\
\hspace{0.20cm} w/o training regularization & 68.5 & 59.7 & 64.1\textsubscript{$\downarrow$1.0} \\
\hspace{0.20cm} w/o \emph{all above} (Baseline) & 50.3 & 52.5 & 51.4\textsubscript{$\downarrow$13.7} \\ 
\hspace{0.20cm} w/o \texttt{Pile-NER} & 69.5 & 60.2 & 64.9\textsubscript{$\downarrow$0.2} \\
        \midrule
GPT-4 & 60.1 & 55.5 & 57.8 \\
\hspace{0.20cm} w/o entity definition std. & 53.0 & 50.6 & 51.8\textsubscript{$\downarrow$6.0} \\
        \bottomrule

    \end{tabular} }
    \caption{Ablation study for both B\textsuperscript{2}NER and GPT-4. F1 scores come from out-of-domain evaluations.} 
    \label{tbl:ablation}
\end{table}

Table \ref{tbl:ablation} details our ablation study on the impact of various components in our approach under OOD evaluations. 
Results on B\textsuperscript{2}NER show significant benefits from entity definition standardization, diversity-aware data pruning, and training regularization.
In contrast, the "w/o \texttt{Pile-NER}" model trained solely with B\textsuperscript{2}NERD data, shows minimal performance regression, indicating the individual effectiveness for our data.
Additionally, adding dataset names ("w/ dataset names") instead of standardizing entity definitions hurts overall performance, confirming that models learn dataset-specific patterns this way (See Appendix \ref{app:case} for a case study).
For GPT-4, skipping the entity definition standardization on test datasets also leads to substantial performance losses, underscoring the overall effectiveness of our entity definition standardization and universal taxonomy on LLMs.

\subsection{In-depth Analysis of Data Pruning}
\label{subsec:analysis}

To better understand the impact of our diversity-aware data pruning method, we compare various sampling strategies and examine data scaling effects. We focus on out-of-domain setting and experiment on Chinese NER data for simplicity.

\begin{figure}[t!]
    \centering
    \includegraphics[width=0.94\columnwidth]{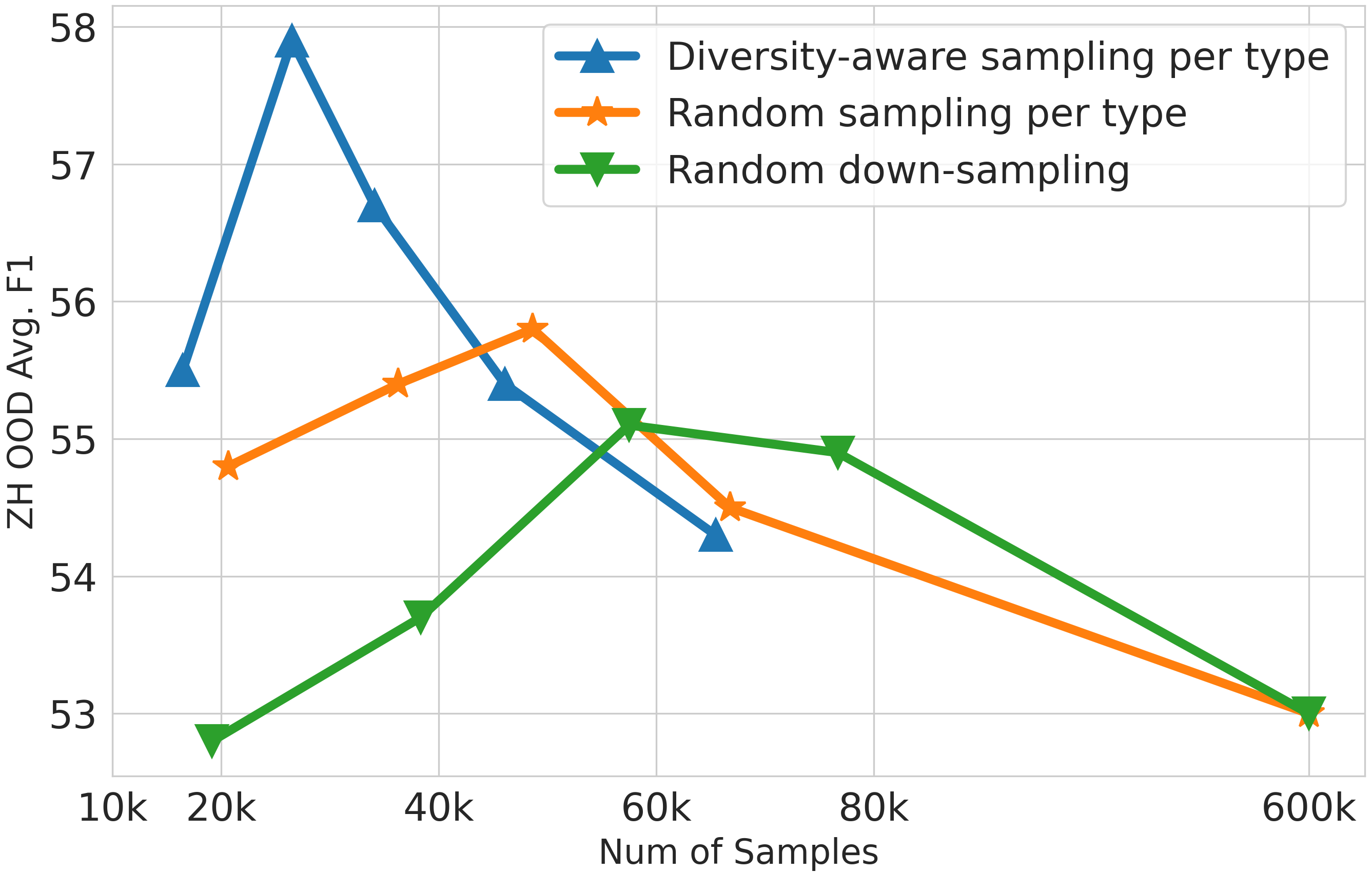}
    \caption{Data scaling results for different sampling methods. Diversity-aware strategy (blue) achieves better performance with fewer samples.}
    \label{fig:lineplot}
\end{figure}

\begin{itemize}[label={}, labelsep=0pt, leftmargin=0pt]
    \item \textbf{Sampling Strategies} \hspace{0.15cm} Beyond our diversity-aware strategy, we evaluate 2 additional methods: 
    1) Random sampling per type, which evenly selects random samples for each entity type, ignoring the semantic diversity. 
    2) Random down-sampling, which selects random samples regardless of entity types. 
    Figure \ref{fig:lineplot} shows that diversity-aware sampling achieves the highest peak performance, while random down-sampling yields the weakest. 
    This highlights the importance of data diversity in tuning LLMs. 
    Variants of diversity-aware sampling strategies are explored in Appendix \ref{app:var}.
    
    \item \textbf{Data Scaling} \hspace{0.15cm} We varied the value of $k$ during data pruning to generate datasets of differing scales for our experiments. 
    The line plots in Figure \ref{fig:lineplot} show that for all sampling strategies, peak performance is achieved with moderate data size; both excess and scarcity of data can hinder model effectiveness. 
    Another clear trend is that \textbf{diversity-aware strategy can achieve better performance with fewer samples}. 
    These results support our assumption that redundant data causes LLMs to over-fit, while curated, diverse data benefits universal generalization. This finding aligns with recent data efficiency research (\citealp{zhou2023lima}; \citealp{liu2024what}; \citealp{ye2024empirical}).    
\end{itemize}

\section{Discussion}
\begin{itemize}[label={}, labelsep=0pt, leftmargin=0pt]

\begin{table*}[t!]
    \centering
    \resizebox{1.98\columnwidth}{!}{
    \begin{tabular}{lll}
        \toprule
        \textbf{Sentence} & \textbf{Prediction} & \textbf{Truth} \\ 
        \midrule
        any restaurants open right now & (\textsc{operating hours}: \textbf{right now}) & (\textsc{operating hours}: \textbf{open right now}) \\
        can i see hamburger restaurants nearby & (\textbf{\textsc{cuisine type}}: hamburger) & (\textbf{\textsc{dish or beverage name}}: hamburger) \\
        \bottomrule
    \end{tabular} }
    \caption{Representative errors from the OOD evaluation on the \texttt{Restaurant} dataset. \textbf{Bold} text highlights the errors. While the model's predictions are reasonable, they do not align with the dataset's specific annotation conventions.}
    \label{tbl:error}
\end{table*}

\item \textbf{Error Analysis} \hspace{0.15cm} 
Despite achieving state-of-the-art generalization by training on our compact and coherent dataset, our best model still produces many errors in current out-of-domain evaluations, with top performances of 72.1 on English and 62.3 on Chinese benchmarks. To guide future work, we analyzed some error cases and identified a key issue: the model struggles to align with the unique annotation standards of each test dataset. For example, Table \ref{tbl:error} presents representative errors from our analysis of the \texttt{Restaurant} dataset, where model predictions are reasonable but do not align with the dataset's specific conventions. These minor, unique inconsistencies are hidden and pervasive across test datasets, making them difficult to address through written descriptions and challenging for out-of-domain models to capture without extensive fine-tuning. We view this type of error as a limitation of current evaluation benchmarks and methods. Therefore, a promising direction for future work is to develop a dedicated benchmark or evaluation method for more accurate assessment of strong Open NER models.

\item \textbf{Flat and Nested NER} \hspace{0.15cm} Our method supports both flat NER and nested NER tasks, but current dataset is mainly developed and tested for flat NER tasks. There are 2 nested NER datasets in our collected datasets: \texttt{ACE2005} and \texttt{GENIA}. We assign distinct entity labels to nested datasets during our standardization process to prevent conflicts with flat datasets, so current models will only extract nested entities for those specific labels. The dataset can be easily reused or extended to train models with better generalization for nested NER tasks by incorporating explicit hints in prompts (e.g., a "nested" tag) or entity labels for nested NER training data.

\end{itemize}

\section{Related Work}
\begin{itemize}[label={}, labelsep=0pt, leftmargin=0pt]

\item \textbf{Instruction Tuning} \hspace{0.15cm} Instruction tuning (\citealp{sanh2022multitask}; \citealp{ouyang2022training}) can boost LLMs' efficacy on unseen tasks via fine-tuning with exemplary natural language instructions. 
Current instruction tuning datasets, constructed from human (\citealp{conover2023free}), LLM (\citealp{wang2022self}; \citealp{xu2023wizardlm}), or existing datasets (\citealp{longpre2023flan}; \citealp{wang2022super}; \citealp{yu2023seqgpt}), mostly prioritize large quantities.
In contrast, recent work (\citealp{zhou2023lima}; \citealp{liu2024what}; \citealp{ye2024empirical}) shows that using fewer but higher quality instruction tuning data could align LLMs better on general tasks. 
Our work, following this direction, focuses on downstream applications like Open NER, where data engineering strategies on how to merge and prune task-specific datasets for efficient instruction tuning are still under-explored.

\item \textbf{Generative NER} \hspace{0.15cm} Numerous attempts have been made to harness LLMs to solve Information Extraction (IE) tasks like NER in a generative paradigm (\citealp{xu2023large}).
Researchers (\citealp{xie2023empirical}; \citealp{wang2023gpt}; \citealp{ashok2023promptner}) leverage LLMs like ChatGPT for NER via in-context learning, which is orthogonal to our approach (see Appendix \ref{app:ann}).
Recent studies use instruction tuning to train custom LLMs with existing datasets (\citealp{wang2023instructuie}; \citealp{gui2024iepile}; \citealp{xiao2023yayi}), but face challenges in Open NER due to dataset inconsistencies and redundancies. 
GoLLIE (\citealp{sainz2023gollie}) trains LLMs to follow detailed code-style annotation guidelines to resolve inconsistent entity definitions, but such guidelines can be difficult to obtain and understand.
Our work takes a different approach that directly clarifies entity ambiguities and restructures existing datasets for optimal LLM learning.
Other studies explore synthetic NER data distilled from LLMs (\citealp{zhou2023universalner}; \citealp{lu2023pivoine}; \citealp{li2024knowcoder}; \citealp{ding2024rethinking}). However, synthetic data often falls short in covering real-world NER tasks comprehensively.

\end{itemize}

\section{Conclusion}
We present \texttt{B\textsuperscript{2}NERD}, a cohesive and compact dataset designed to enhance LLMs for Open NER.
Refined from 54 datasets through entity definition standardization and diversity-aware data pruning, \texttt{B\textsuperscript{2}NERD} addresses inconsistencies and redundancies in existing datasets, enabling LLMs to learn a universal entity taxonomy beyond data boundaries.
Models trained on \texttt{B\textsuperscript{2}NERD} outperform GPT-4 and previous methods in out-of-domain evaluations across various datasets and languages.
We will share our recipe and data to support further research.

\section*{Limitations}
While our work contributes to stronger LLMs for the Open NER task, it has following limitations:

\begin{itemize}
    \item \textbf{Benchmarks}: Current out-of-domain evaluation is mainly performed by holding out certain datasets from existing ones. However, these test datasets may contain unique annotation standards that can't be learned via OOD generalization and may suffer from data contamination. Based on our error analysis for our 20B model, we are concerned that the ceiling for these datasets' OOD evaluation may soon be approached. A dedicated comprehensive benchmark for Open NER evaluation may be necessary in the near future.
    \item \textbf{Diversity Measure}: In our existing data pruning strategy, we evaluate the diversity of entity types and text semantics independently. Semantic diversity is assessed within the context of each entity type. Yet, a more inclusive measure could be developed to simultaneously compare annotations and text in pairs of samples. Such an approach might enable globally optimal data selection, encapsulating more information with fewer samples and providing insights on what kind of data are best for task-focused instruction tuning. 
\end{itemize}

\section*{Acknowledgements}
The authors wish to thank the anonymous reviewers for their helpful comments. This work was partially funded by National Natural Science Foundation of China (No. 62376061, 62206057, 62076069), the Major Key Project of PCL under Grant PCL2024A06, Shanghai Rising-Star Program (23QA1400200), Natural Science Foundation of Shanghai (23ZR1403500), Program of Shanghai Academic Research Leader under grant 22XD1401100.

\bibliography{custom}

\clearpage

\appendix

\section{Implementation Details}
\label{app:imple}

\subsection{Details of Data Collection and Cleaning}
\label{app:dcc}

We spend non-trivial effort on data collection and data cleaning for Chinese NER data by following major steps. \textbf{Data collection}, after an extensive search, we initially identify 35 publicly available Chinese NER datasets, about half of which is never used by previous works. \textbf{Deduplication}, We remove those datasets that have highly duplicate data with others. For example, \texttt{peopledaily1998} dataset is actually part of \texttt{MSRA} dataset. \textbf{Annotation quality screening}, as many datasets didn't share details on their labelling process, we manually re-evaluate their annotation consistency at dataset level. Datasets with low internal consistency are excluded. \textbf{Label name translation}, many datasets use English symbols and Arabic numbers as entity type name, such as "PER", "HCCX", "1". To help LLM understand the entity type together with input Chinese text, we translate all type names into natural language in Chinese. We prompt GPT4 to help the translation. These label names will be further standardized in Section \ref{EDS}. 

\subsection{Details of Automatic Dataset Conflict Detection}
\label{app:autod}
\begin{figure}[t!]
    \centering
    \includegraphics[width=1.0\columnwidth]{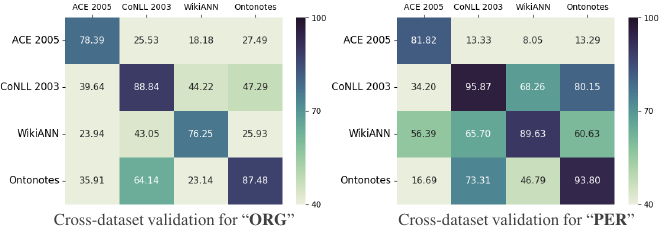}
    \caption{More results from model-based cross validation on \textsc{PER} and \textsc{ORG} among 4 datasets. The horizontal axis represents testing data and vertical represents training data.}
    \label{fig:mcon}
\end{figure}

For model-based cross validation, we implement a BERT-CRF model\footnote{\url{https://github.com/lonePatient/BERT-NER-Pytorch}} to learn from one dataset and infer on others with similar entity types. Figure \ref{fig:mcon} shows more results from this cross validation. 

To better understand the conflicts reflected in model-based cross-validation results, we examine the surprisingly low F1 scores for the \textsc{LOC} entity across two popular datasets, \texttt{CoNLL} and \texttt{OntoNotes} (Figure \ref{fig:conflicts}). 
Our analysis indicates that these discrepancies indeed arise from differences in entity definitions. 
\texttt{OntoNotes} defines \textsc{LOC} as "non-GPE locations, mountain ranges, and bodies of water" with a separate \textsc{GPE} type for geo-political entities\footnote{\url{https://catalog.ldc.upenn.edu/docs/LDC2011T03/OntoNotes-Release-4.0.pdf}}, while \texttt{CoNLL} includes \textsc{GPE} within \textsc{LOC}. 
Additionally, many \textsc{LOC} mentions in \texttt{CoNLL}, such as "New Zealand" and "Minnesota," are actually \textsc{GPE} entities, due to the dataset's news source. 
Thus, as shown in Table \ref{tbl:detail_cv}, the model trained on \texttt{CoNLL} tends to mislabel \texttt{OntoNotes}'s \textsc{GPE} entities as \textsc{LOC}, resulting in low accuracy. 
Conversely, the reverse model annotates many \textsc{LOC} entities in \texttt{CoNLL} as \textsc{GPE}, leading to low recall.

\begin{table}[t!]
    \centering
    \resizebox{0.80\columnwidth}{!}{
    \begin{tabular}{lccc}
        \toprule
        \textbf{Train $\rightarrow$ Test} & \textbf{Precision} & \textbf{Recall} & \textbf{F1} \\ 
        \midrule
        CoNLL $\rightarrow$ OntoNotes & 4.64 & 65.05 & 8.66 \\
        OntoNotes $\rightarrow$ CoNLL & 64.52 & 1.20 & 2.35 \\
        \bottomrule
    \end{tabular} }
    \caption{Detailed metrics for the model-based cross validation between \texttt{CoNLL} and \texttt{OntoNotes} for \textsc{LOC} entity.}
    \label{tbl:detail_cv}
\end{table}

For rule-based screening, we employ detailed rules to ensure accurate detection. We screen for cases where 1) two datasets share one entity label (e.g., \textsc{location}); 2) the same entity mention appeared in the samples of both datasets (e.g., "Belgium"); 3) this mention receives inconsistent recognition results in two datasets (e.g., \textsc{location} v.s. \textsc{geo-political entity}). We also exclude cases where current mention is part of another extracted entity, as it's reasonable for flat NER datasets not to extract this mention. After screening, we classify the error types of inconsistent cases for each pair of conflicting entities and list significant ones for future processing. Error types include wrong categories, not extracted and partially extracted. 

\subsection{Details of Entity Definition Standardization}
\label{app:EDS}

We publicly recruit 4 college students with prior NER annotation experience as human experts to delineate entity definition differences, re-assign label names, and write annotation guidelines (for the experiments in Appendix \ref{app:ann}). These experts, including one with a biomedical background, are selected for their strong understanding of entity definitions and annotation guidelines. Since the task involves providing proper entity label names for existing entity types on a dataset basis rather than re-labeling individual samples, their prior experience and knowledge enable them to perform the task with sufficient expertise after some training. Compensation for the annotators is provided on an hourly basis and exceeds local standards, reflecting the effort and expertise required for this task.

\begin{itemize}[label={}, labelsep=0pt, leftmargin=0pt]

\item \textbf{Provided Resources and Instructions} \hspace{0.15cm} 
The annotators are equipped with several resources to aid their work, including conflict detection results (as described in Appendix \ref{app:autod}), which show conflict statistics, error types of conflicting entity definitions, and examples of the conflicts. They also receive raw datasets, sample annotations from these datasets, and data summaries, such as the most frequent entity mentions. Additionally, our naming principles with examples (outlined in Section \ref{EDS}) and supplementary materials, such as the ACE annotation guidelines, are provided. Based on this information, annotators are tasked with screening each entity type, writing proper label names, and justifying their choices based on the naming principles and previous annotation guidelines.

\item \textbf{LLMs as an Auxillary Tool} \hspace{0.15cm}
LLMs are also used to accelerate the process. Annotators receive a standard prompt instructing the LLM to suggest proper label names based on entity annotation examples and our naming principles. While annotators are welcome to use their own prompts to explore additional insights, the final label names are determined through careful human consideration. The LLM suggestions primarily act as a reference, ensuring that human expertise remains central to the decision-making process.

\item \textbf{Time Allocation and Workflow} \hspace{0.15cm}
Each expert undergoes 3 hours of initial training to familiarize themselves with the task, resources, and naming principles. They then spend 12 hours screening and renaming approximately 120 assigned entity types each, based on detected conflicts and guidelines. Following this, 5 hours are allocated for group discussions, during which the experts collectively review and check the consistency of all entity labels, finalize names, and resolve any ambiguities. In total, the four annotators collectively spend approximately 80 hours on this process.

\item \textbf{Attention Checks} \hspace{0.15cm}
During the final discussion phase, the authors review the annotators' work for quality by examining the provided entity labels and the justifications for each. Ill-defined entity types are also eliminated as part of this process. Common entities encountered by most annotators serve as attention checks to ensure accuracy and consistency. The high-quality work provided by the annotators, combined with the rigorous group discussions, ensures that the final universal taxonomy is both accurate and consistent. 
\end{itemize}

\subsection{Details of Training Regularization}
\label{app:trainreg}

We observe that the co-occurrence of entity label options in instructions is an obvious pattern for samples coming from same original dataset. For example, data from \texttt{Taobao} dataset all ask to recognize \textsc{Product Name} and \textsc{Brand} in their instructions. LLMs may just memorize this dataset-specific co-occurrence pattern without understanding the given label names. To alleviate this, we introduce regularization methods, including: \textbf{Dynamic label set.} Instead of asking LLM to recognize a static set of labels, we mention random entity types in random order in instructions for less co-occurrence patterns. Labels that current sample contains still remain in instructions to assure the answer is still correct. \textbf{Random label dropout.} We randomly neglect some entity types in both the instruction and answer of a sample. This can force LLM to focus on target label names in instructions when generating answers.

\section{More Experiments and Studies}
\label{app:moreexp}

\begin{table}[t!]
    \centering
    \resizebox{1.0\columnwidth}{!}{
\begin{tabular}{c|ccc|c}
\toprule
Dataset & BERT-base & InstructUIE-11B & UniNER-7B & B\textsuperscript{2}NER-7B \\
\midrule
ACE05 & \textbf{87.30} & 79.94 & 86.69 & 83.04 \\
AnatEM & 85.82 & 88.52 & 88.65 & \textbf{89.18}\\
bc2gm & 80.90 & 80.69 & \textbf{82.42} & 81.95\\
bc4chemd & 86.72 & 87.62 & \textbf{89.21} & 88.96\\
bc5cdr & 85.28 & 89.02 & \textbf{89.34} & 88.52\\
Broad Twitter & 58.61 & 80.27 & 81.25 & \textbf{82.16}\\
CoNLL03 & 92.40 & 91.53 & \textbf{93.30} & 92.56\\
FabNER & 64.20 & 78.38 & \textbf{81.87} & 78.82\\
FindVehicle & 87.13 & 87.56 & \textbf{98.30} & 97.89\\
GENIA & 73.3 & 75.71 & \textbf{77.54} & 76.43\\
HarveyNER & \textbf{82.26} & 74.69 & 74.21 & 73.67\\
MIT Movie & 88.78 & 89.58 & 90.17 & \textbf{90.78}\\
MIT Restaurant & 81.02 & 82.59 & 82.35 & \textbf{83.71}\\
MultiNERD & 91.25 & 90.26 & 93.73 & \textbf{93.98}\\
ncbi & 80.20 & 86.21 & \textbf{86.96} & 84.83\\
OntoNotes & \textbf{91.11} & 88.64 & 89.91 & 84.31 \\
PolyglotNER & \textbf{75.65} & 53.31 & 65.67 & 61.96\\
TweetNER7 & 56.49 & 65.95 & 65.77 & \textbf{66.26}\\
WikiANN & 70.60 & 64.47 & 84.91 & \textbf{85.07}\\
wikiNeural & 82.78 & 88.27 & \textbf{93.28} & 93.01\\ \midrule
Avg & 80.09 & 81.16 & \textbf{84.78} & 83.85 \\
\bottomrule
\end{tabular} }
    \caption{Full results of in-domain supervised evaluation on English NER.} 
    \label{tbl:supen}
\end{table}

\begin{table}[t!]
    \centering
    \resizebox{1.0\columnwidth}{!}{
\begin{tabular}{c|ccc|c}
\toprule
Dataset & BERT-base & YAYI-UIE & IEPile & B\textsuperscript{2}NER-7B \\
\midrule
CCKS2017     & 92.68 & 90.73 & - & \textbf{94.93} \\
MSRA         & \textbf{96.72} & 95.57 & 87.99 & 92.22\\
Multiconer22 & 69.78 & - & - & \textbf{71.53}\\
Multiconer23 & 66.98 & - & - & \textbf{69.56}\\
resume       & \textbf{96.01} & - & 93.92 & 95.90\\
Youku        & 86.26 & - & - & \textbf{86.50}\\
\midrule
Avg          & 84.74 & - & - & \textbf{85.11} \\
\bottomrule
\end{tabular} }
    \caption{Full results of in-domain supervised evaluation on Chinese NER.} 
    \label{tbl:supzh}
\end{table}

\subsection{In-domain Supervised Evaluation Results}
\label{app:idmore}

Table \ref{tbl:supen} shows the full results for in-domain supervised evaluation on English NER. Though trailing UniNER-7B by 1 point on average, B\textsuperscript{2}NER-7B achieves best results in 7 out of 20 datasets.

Table \ref{tbl:supzh} shows the full results for in-domain supervised evaluation on Chinese NER. We do not use the complete Chinese training datasets for in-domain supervised evaluation because some datasets lack high-quality test sets in their original splits. Our B\textsuperscript{2}NER-7B model achieves the best performance on 4 out of 6 datasets and surpasses BERT-based models on average.

\subsection{Results of Different Backbones}
\label{app:backb}

We investigate the effectiveness of our approach and the \texttt{B\textsuperscript{2}NERD} dataset on different backbone models. 
In addition to InternLM2-7B, we further fine-tune our dataset on Baichuan2-7B (\citealp{baichuan2023baichuan2}) and Llama2-7B (\citealp{touvron2023llama}). 

\begin{table}[t!]
    \centering
    \resizebox{0.80\columnwidth}{!}{
    \begin{tabular}{lcc|c}
        \toprule
        \textbf{Model} & \textbf{EN} & \textbf{ZH} & \textbf{OOD Avg.} \\ 
        \midrule

B\textsuperscript{2}NER-InternLM2-7B & \textbf{69.8} & \textbf{60.3} & \textbf{65.1} \\
B\textsuperscript{2}NER-Baichuan2-7B & 67.9 & 57.9 & 62.9 \\
B\textsuperscript{2}NER-Llama2-7B    & 66.7 & 42.6 & 54.7 \\
GPT-4                                & 60.1 & 55.5 & 57.8 \\
        \bottomrule

    \end{tabular} }
    \caption{Out-of-domain evaluation results of fine-tuning different backbone models with our \texttt{B\textsuperscript{2}NERD} dataset. } 
    \label{tbl:backb}
\end{table}

As shown in Table \ref{tbl:backb}, all models achieve superior out-of-domain performance over GPT-4 except B\textsuperscript{2}NER-Llama2-7B, which trails behind on Chinese NER. 
B\textsuperscript{2}NER-InternLM2-7B achieves best overall performance.

\subsection{Compatibility with In-Context Learning}
\label{app:ann}

\begin{table}[t!]
    \centering
    \resizebox{0.90\columnwidth}{!}{
    \begin{tabular}{lc|cc}
        \toprule
        \textbf{Model} & \textbf{0-shot} & \textbf{w/ guidelines} & \textbf{3-shot}  \\
        \midrule

Baseline (only Chinese)                & 52.5 & 56.5 & 58.4 \\
B\textsuperscript{2}NER (only Chinese) & 57.9 & 62.3 & \textbf{62.6} \\
        \bottomrule
    \end{tabular} }
    \caption{Study on additional in-context learning methods with annotation guidelines or few-shot examples. "0-shot" denotes our out-of-domain evaluation using zero-shot instruction template.}
    \label{tbl:ann}
\end{table}

As our method is orthogonal to other in-context learning approaches, such as adding annotation guidelines (\citealp{sainz2023gollie}) and few-shot examples (\citealp{wang2023gpt}), we explore their combined performance. 
Focusing on Chinese NER, we invite experts to write guidelines for Chinese datasets in \texttt{B\textsuperscript{2}NERD}. 
Models are trained with these guidelines or few-shot examples using instruction templates in Appendix \ref{app:it}. 

Results in Table \ref{tbl:ann} show that our "B\textsuperscript{2}NER (only Chinese)" model can be further improved with in-context learning, demonstrating the compatibility. Notably, the Baseline model with guidelines still fails to outperform B\textsuperscript{2}NER (56.5 < 57.9), highlighting our approach's superior effectiveness over additional guidelines. 

\subsection{Variations of Diversity-aware Sampling Strategy}
\label{app:var}
We experimented with other diversity-aware sampling strategies during data pruning. 
One alternative is the "threshold filter per type," which uses a hard semantic distance threshold instead of probabilistic sampling for selecting samples for each entity type's pool. 
We also tried different offsets $b$ for the semantic distance measure, as introduced in Section \ref{DDS}. 
A higher offset imposes a stricter semantic distance requirement, resulting in more diverse semantics.

\begin{figure}[t!]
    \centering
    \includegraphics[width=1.0\columnwidth]{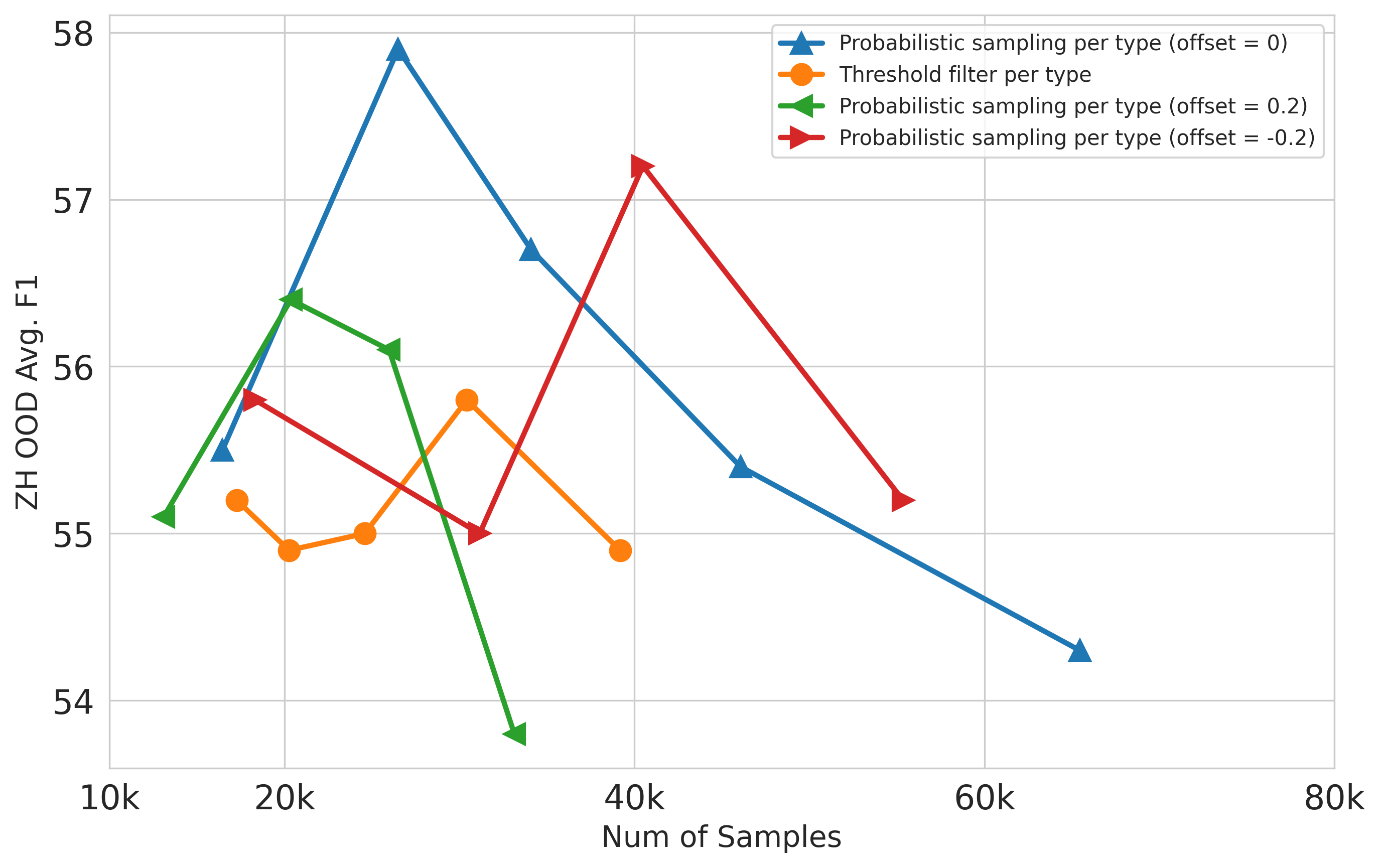}
    \caption{Data scaling results for variants of diversity-aware sampling strategy. Higher offset means more strict semantic distance requirement.}
    \label{fig:lineplot2}
\end{figure}

Figure \ref{fig:lineplot2} shows that an offset of 0 achieves the best peak performance. Additionally, higher offsets reach peak performance with fewer samples, indicating that greater semantic diversity can compress information into a smaller dataset.

\subsection{Case Study on Dataset-Specific v.s. General Patterns}
\label{app:case}
\begin{figure}[t!]
    \centering
    \includegraphics[width=1.0\columnwidth]{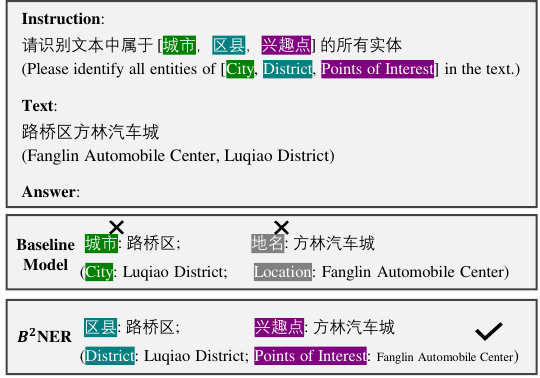}
    \caption{An out-of-domain NER example on the dataset of \texttt{CCKS2021address}, where baseline model displays dataset-specific patterns.}
    \label{fig:example}
\end{figure}

Figure \ref{fig:example} shows an out-of-domain NER example on \texttt{CCKS2021address} dataset. 
The baseline model trained with raw inconsistent datasets produces incorrect and out-of-scope entity types, reflecting its learning of dataset-specific patterns from prior City-Location style data. 
In contrast, Our B\textsuperscript{2}NER delivers accurate results for this unseen task, owing to our instruction tuning method that transcends data boundaries.

\subsection{Experiments on Two-stage Training}
Previous studies, such as \citealp{ding2024rethinking}, use a two-stage training strategy, starting with general \texttt{Pile-NER} data followed by supervised datasets. We test this approach on the OOD English NER benchmark.

\begin{table}[h!]
    \centering
    \resizebox{0.9\columnwidth}{!}{
    \begin{tabular}{lcc}
        \toprule
        \textbf{} &  \textbf{Two-Stage Training} & \textbf{Mixed Training} \\
        \midrule

OOD Avg. F1 & 69.7 & 69.8 \\
        \bottomrule
    \end{tabular} }
    \caption{Comparison between two-stage training and mixed training strategies}
    \label{tbl:2stage}
\end{table}

In Table \ref{tbl:2stage}, "Two-Stage Training" involves sequential training with \texttt{Pile-NER} and \texttt{B\textsuperscript{2}NERD}, while 'Mixed Training' refers to our strategy of training them simultaneously. The similar performance of both approaches suggests that each is equally effective in our LoRA training scenario.

\section{Training Details}

\subsection{Hyper-parameters}
\label{app:train}

As explained in Section \ref{subsec:setup}, we use InternLM2 (\citealp{cai2024internlm2}) as the backbone model and fine-tune it with LoRA (\citealp{hu2021lora}) to derive all the models.
Although we also experiment with other bilingual backbone LLMs in Appendix \ref{app:backb}, InternLM2 demonstrates better overall performance.

For our main out-of-domain experiments, we apply LoRA to "wqkv" target modules, setting $r$ to 32 and the dropout rate to 0.05. 
Preliminary comparisons between full-parameter tuning and LoRA show that LoRA provides better and more stable results.
During training, we use 3e-4 learning rate with warmup ratio of 0.02 and a cosine scheduler. 
DeepSpeed (\citealp{rasley2020deepspeed}) Stage 2 is adopted for memory optimization. 
The main model, B\textsuperscript{2}NER, is trained with a batch size of 128. 
For datasets of varying sizes, we experimented with different batch sizes to find the most effective configuration. 
The maximum context length of our LLM is set to 4096 tokens.
Training and inference are done on one 8$\times$ Nvidia-A100-40G node, with a single run of 5 epochs taking about 8 hours.

For in-domain supervised experiments, we use full-parameter tuning with a learning rate of 2e-5 and disable the training regularization methods in Appendix \ref{app:trainreg}. All other settings match the out-of-domain experiments.

\subsection{Instruction Templates}
\label{app:it}

\begin{figure}[t!]
    \centering
    \includegraphics[width=0.9\columnwidth]{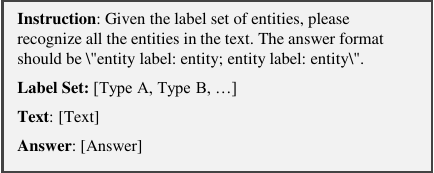}
    \caption{Instruction template for our main experiments.}
    \label{fig:0shot}
\end{figure}

\begin{figure}[t!]
    \centering
    \includegraphics[width=0.9\columnwidth]{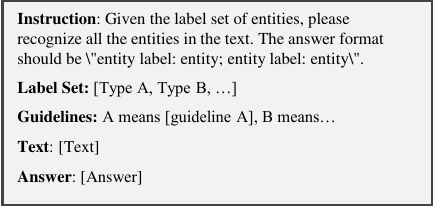}
    \caption{Instruction template with annotation guidelines for experiments in Appendix \ref{app:ann}.}
    \label{fig:guideline}
\end{figure}

\begin{figure}[t!]
    \centering
    \includegraphics[width=0.9\columnwidth]{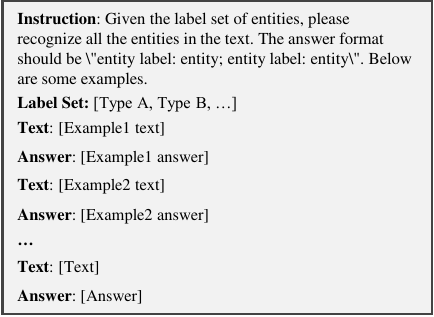}
    \caption{Few-shot instruction template for experiments in Appendix \ref{app:ann}.}
    \label{fig:fewshot}
\end{figure}

Figure \ref{fig:0shot} displays the instruction template used in our main experiments. For the study in Appendix \ref{app:ann}, Figure \ref{fig:guideline} shows our instruction template when annotation guidelines are available; Figure \ref{fig:fewshot} shows the template for our few-shot experiments.

\section{Dataset and Taxonomy}
\label{app:dst}

\subsection{Dataset Statistics}
\label{app:fullstats}
Table \ref{tbl:dataset-en} shows the statistics of all English datasets inside \texttt{B\textsuperscript{2}NERD}. 
Table \ref{tbl:dataset} shows the statistics of all Chinese datasets.

All datasets are flat NER datasets, except for two nested NER datasets: \texttt{ACE2005} and \texttt{GENIA}.
Note that for \texttt{CAIL2021}, we randomly split 20\% samples as test set for evaluation. 
Other datasets all inherit original train and test splits.

\subsection{Full NER Taxonomy}
\label{app:fulltax}
Table \ref{tbl:en-taxonomy} shows the full NER taxonomy of English entities in \texttt{B\textsuperscript{2}NERD}. 
Table \ref{tbl:zh-taxonomy2} shows the full NER taxonomy of Chinese entities.

\clearpage

\begin{table*}
        \resizebox{\textwidth}{!}{
        \begin{tabular}{cc|c|cc|ccc}
            \Xhline{2pt}
             \rule{0pt}{2.4ex} 
            \textbf{Dataset} & \textbf{Types} & \textbf{Major Domain} & \textbf{Train} & \textbf{Test} & \textbf{Pruned Train Num} & \textbf{Raw Train Num} & \textbf{Test Num}\\
    
            \Xhline{1pt}

    ACE2004(\citealp{mitchell2005ace}) & 7 & News & $\checkmark$ & & 1193 & 6177 & 812 \\
    ACE2005(\citealp{walker2006ace}) & 7 & News  & $\checkmark$ & $\checkmark$ & 1433 & 7134 & 1050 \\
    AnatEM(\citealp{pyysalo2014anatomical}) & 1 & Biomedical & $\checkmark$ &$\checkmark$ & 480 & 5667 & 3758\\
    bc2gm(\citealp{kocaman2021biomedical}) & 1 & Biomedical & $\checkmark$ & $\checkmark$ & 480 & 12392 & 4977 \\
    bc4chemd(\citealp{kocaman2021biomedical}) & 1 & Biomedical & $\checkmark$ & $\checkmark$ & 480 & 30488 & 26204 \\
    bc5cdr(\citealp{zhang2022optimizing}) & 2 & Biomedical & $\checkmark$ & $\checkmark$ & 592 & 4545 & 4788 \\
    Broad Tweet Corpus(\citealp{derczynski2016broad}) & 3 & Social Media & $\checkmark$ & $\checkmark$ & 855 & 5324 & 2000\\
    CoNLL 2003(\citealp{sang2003introduction}) & 4 & News & $\checkmark$ & $\checkmark$ & 1069 & 12613 & 3184 \\
    FabNER(\citealp{kumar2022fabner}) & 12 & Scientific & $\checkmark$ & $\checkmark$ & 1741 & 9421 & 2064 \\
    FindVehicle(\citealp{guan2024findvehicle}) & 21 & Traffic & $\checkmark$ & $\checkmark$ & 2591 & 21547 & 20769\\
    GENIA\_NER(\citealp{kim2003genia}) & 5 & Biomedical & $\checkmark$ & $\checkmark$ & 1281 & 14966 & 1850\\
    HarveyNER(\citealp{chen2022crossroads}) & 4 & Social Media & $\checkmark$ & $\checkmark$ & 405 & 3553 & 1260\\
    MultiNERD(\citealp{tedeschi2022multinerd}) & 16 & Wikipedia & $\checkmark$ & $\checkmark$ & 4659 & 130623 & 9994\\
    ncbi(\citealp{dougan2014ncbi}) & 1 & Biomedical & $\checkmark$ & $\checkmark$ & 480 & 5432 & 940\\
    Ontonotes(\citealp{hovy2006ontonotes}) & 18 & General & $\checkmark$ & $\checkmark$ & 4343 & 54994 & 7782\\
    PolygoltNER(\citealp{polyglotner}) & 3 & Wikipedia & $\checkmark$ & $\checkmark$ & 0 & 393941 & 10000\\
    TweetNER7(\citealp{ushio2022named}) & 7 & Social Media & $\checkmark$ & $\checkmark$ & 1325 & 7111 & 576\\
    Wikiann en(\citealp{rahimi-etal-2019-massively}) & 3 & Wikipedia & $\checkmark$ & $\checkmark$ & 856 & 20000 & 10000\\
    WikiNeutral(\citealp{tedeschi-etal-2021-wikineural-combined}) & 3 & Wikipedia & $\checkmark$ & $\checkmark$ & 1140 & 92720 & 11597\\
    \Xhline{1pt}
    CrossNER\_AI(\citealp{liu2021crossner}) & 14 & AI & & $\checkmark$ & / & / & 431\\
    CrossNER\_literature(\citealp{liu2021crossner}) & 12 & Literary  & & $\checkmark$ & / & / & 416\\
    CrossNER\_music(\citealp{liu2021crossner}) & 13 & Musical & & $\checkmark$ & / & / & 465\\
    CrossNER\_politics(\citealp{liu2021crossner}) & 9 & Political & & $\checkmark$ & / & / & 650\\
    CrossNER\_science(\citealp{liu2021crossner}) & 17 & Scientific & & $\checkmark$ & / & / & 543\\
    mit-movie(\citealp{liu2013asgard}) & 12 & Social Media & & $\checkmark$ & / & 9707 & 2441\\
    mit-restaurant(\citealp{liu2013asgard}) &8 &  Social Media & & $\checkmark$ & / & 7658 & 1520\\
    
            \Xhline{2pt}
        \end{tabular}}
        \captionof{table}{Statistics of English datasets in \texttt{B\textsuperscript{2}NERD}.}
        \label{tbl:dataset-en}
    \vspace{6cm}
\end{table*}

\clearpage

\begin{minipage}{\textwidth}
        \centering
        \resizebox{\textwidth}{!}{
        \begin{tabular}{cc|cc|cc|ccc}
            \Xhline{2pt}
             \rule{0pt}{2.4ex} 
            \textbf{Dataset} & \textbf{Types} & \textbf{Major Domain} & \textbf{Source} & \textbf{Train} & \textbf{Test} & \textbf{Pruned Train Num} & \textbf{Raw Train Num} & \textbf{Test Num}\\
    
            \Xhline{1pt}
    
    Bank\footnote{\url{https://www.heywhale.com/mw/dataset/617969ec768f3b0017862990/file}} & 4 & Social Media & Online Forum & $\checkmark$ &  & 1224 & 10000 & / \\
    Boson\footnote{Boson,Peopledaily2014 and Finance-sina datasets are available at \url{https://github.com/liucongg/NLPDataSet}} & 6 & News & News & $\checkmark$ &   & 876 & 2000 & /\\	
    CCKS2017\footnote{\url{https://www.biendata.xyz/competition/CCKS2017_2/}} & 5 & Biomedical & Medical Records & $\checkmark$ & $\checkmark$ & 505 & 2006 & 223 \\
    CCKS2018\footnote{\url{https://www.biendata.xyz/competition/CCKS2018_1/}} & 5 & Biomedical & Medical Records & $\checkmark$ &  & 212 & 797	 & / \\
    CCKS2019\footnote{\url{https://www.biendata.xyz/competition/ccks_2019_1/}} & 6 & Biomedical & Medical Records & $\checkmark$ &  & 422 & 1379 & / \\	
    CCKS2020\footnote{\url{https://www.biendata.xyz/competition/ccks_2020_2_1/}} & 6 & Biomedical & Medical Records & $\checkmark$ &  & 461 & 1450 & / \\
    Chinese Literature(\citealp{xu2017discourse}) & 10 & Literature & Literature & $\checkmark$ &  & 1675 & 24165	 & 2837\\
    Finance-sina & 4 & News & News & $\checkmark$ &  & 664 & 1579 & / \\
    GAIIC2022\footnote{\url{https://www.heywhale.com/home/competition/620b34ed28270b0017b823ad/content/2}} & 50 & Social Media & Product Titles & $\checkmark$ &  & 2061 & 6776 & / \\
    IMCS21\footnote{\url{http://www.fudan-disc.com/sharedtask/imcs21/index.html}} & 5 & Biomedical & Medical Conversations & $\checkmark$ &  & 1773 & 98529 & 	32935\\
    MSRA(\citealp{levow2006third}) & 3 & News & News & $\checkmark$ & $\checkmark$ & 867 & 45000 & 3442\\
    Multiconer22(\citealp{malmasi-etal-2022-multiconer}) & 6 & Mixed & Wikipedia+Question+Queries & $\checkmark$ & $\checkmark$ & 1889 & 15300 & 151661\\
    Multiconer23(\citealp{fetahu2023semeval}) & 33 & Mixed & Wikipedia+Question+Queries & $\checkmark$ & $\checkmark$ & 4839 & 9759 & 20265\\
    NLPCC2018\footnote{\url{http://tcci.ccf.org.cn/conference/2018/taskdata.php}} & 15 & Voice & Voice Assistants & $\checkmark$ &  & 1754 & 21352 & / \\
    Peopledaily2014 & 4 & News & News & $\checkmark$ &  & 1084 & 286268 & /\\
    Resume(\citealp{zhang2018chinese}) & 8 & News & Resume & $\checkmark$ & $\checkmark$ & 986 & 3821 & 477\\
    Ruijinmcc diabetes\footnote{\url{https://tianchi.aliyun.com/markets/tianchi/ruijin##guid-03}} & 15 & Biomedical & Medical Books + Papers & $\checkmark$ &  & 2680 & 24157 & 2682\\
    Taobao(\citealp{jie2019better}) & 4 & Social Media & Product Titles & $\checkmark$ &  & 982 & 6000 & 1000\\
    Wanchuang\footnote{\url{https://tianchi.aliyun.com/competition/entrance/531824/information}} & 13 & Biomedical & Drug Description & $\checkmark$ &  & 506 & 1255& /\\
    Microbiota\footnote{\url{https://www.heywhale.com/mw/dataset/609a27c5f29cea00179233f3/file}} & 7 & Biomedical & Medical News & $\checkmark$ &  & 71 & 99& /\\
    Youku(\citealp{jie2019better}) & 3 & Social Media & Video Titles & $\checkmark$ & $\checkmark$ & 972 & 8001 & 1001\\
    FNED\footnote{\url{https://www.datafountain.cn/competitions/561/datasets}} & 7 & News & News &  &  & 0 & 10500 & /\\
    
    Military-ner\footnote{\url{https://www.biendata.xyz/competition/ccks_2020_8/}} & 3 & Military & Military &  &  & 0 & 320 & 80\\
            \Xhline{1pt}
    CAIL2021\footnote{\url{http://cail.cipsc.org.cn/task_summit.html?raceID=7&cail_tag=2021}} & 10 & Law & Case description &   & $\checkmark$ & / &4197	 &  1050\\
    CCKS2021address\footnote{\url{https://tianchi.aliyun.com/competition/entrance/531900/introduction}} & 17 & Address & Address &   & $\checkmark$ & / &8856 & 1970\\
    CLUENER(\citealp{xu2020cluener2020}) & 10 & News & News &   & $\checkmark$ & / &10748 & 1343 \\
    CBLUE(\citealp{zhang-etal-2022-cblue}) & 9 & Biomedical & Medical books &   & $\checkmark$ & / &15000 & 4999\\
    Math-high\footnote{\url{https://blog.csdn.net/qq_36426650/article/details/87719204}} & 2 & Math & Math books &   & $\checkmark$ & / &1953 & 279\\
    Weibo(\citealp{peng2015named}) & 8 & Social media	& Social media &   & $\checkmark$ & / &1350 & 270\\
    Zh-Ontonotes(\citealp{weischedel2011ontonotes}) & 4 & News & News &   & $\checkmark$ & / &15724 & 	4346\\
            \Xhline{2pt}
        \end{tabular}}
        \captionof{table}{Statistics of Chinese datasets in \texttt{B\textsuperscript{2}NERD}.}
        \label{tbl:dataset}
    \vspace{7cm}
\end{minipage}

\begin{CJK}{UTF8}{gbsn}
\begin{table*}
    \centering
    \resizebox{\textwidth}{!}{
    \begin{tabular}{c|c|c|c}
        \Xhline{2pt}
         \rule{0pt}{2.4ex} 
        \textbf{person} & \textbf{organization} & \textbf{life} & \textbf{education}\\

        \Xhline{1pt}
        \rule{0pt}{2.6ex} 
general mention of person&general mention of organization&movie actor&AI algorithm\\
mythical figure&organization&movie age rating&academic conference\\
person&organization (without political group)&movie character&academic discipline\\
person -> writer&organization -> university&movie director&academic journal\\
person -> musical artist&organization -> band&movie genre&application domain\\
person -> others&organization -> corporation&movie plot&scientific theory\\
person -> politician&organization -> group or band&movie quality rating or descriptor&experiment metrics\\
person -> researcher&organization -> others&movie song mention&research field\\
person -> scientist&organization -> political party&movie title&research task\\
&&movie trailer or preview term&\\
&&restaurant amenity service&\\
&&restaurant name&\\
&&restaurant quality descriptor&\\
&&cuisine type&\\
&&&\\

        \Xhline{2pt}
        \rule{0pt}{2.4ex} 
        \textbf{location} & \textbf{object}& \textbf{biomedical} & \textbf{others}\\
        \Xhline{1pt}
\rule{0pt}{2.6ex} 
country&animal&DNA&award\\
exact location&astronomical object&RNA&review related term\\
general mention of geo-political entity&orientation of vehicle&anatomy&dish or beverage name\\
general mention of location -> facility&brand of vehicle&biological molecules&else\\
general mention of location -> others&color of vehicle&biomedical term&engineering material\\
geo-political entity&food items&cell line&language\\
geographical area&general mention of vehicle&cell type&legal document\\
location&general mention of weapon&chemical&literary genre type\\
road&machine or equipment&chemical compound&manufacturing concept or principle\\
location (without country)&musical instrument&chemical element&manufacturing process\\
location (without geo-political entity)&technological instrument&disease&manufacturing standard\\
location -> facility&plant&disease name&manufacturing technology\\
nationalities or political group&position of vehicle&enzyme name&mechanical property\\
proximity or location description&product name&gene&miscellaneous\\
river&product name -> vintage car&microorganism&music genre\\
&product name -> MPV&protein name&programming language\\
&product name -> SUV&&process evaluation technique\\
&product name -> bus&&\\
&product name -> coupe&&\\
&product name -> estate car&&\\
&product name -> hatchback&&\\
&product name -> motorcycle&&\\
&product name -> roadster&&\\
&product name -> sedan&&\\
&product name -> sports car&&\\
&product name -> truck&&\\
&product name -> van&&\\
&product name -> vehicle&&\\
&vehicle type&&\\
&vehicle velocity&&\\
&vehicle model&&\\
&vehicle range&&\\
&&&\\
\Xhline{2pt}
        \rule{0pt}{2.4ex} 
        \textbf{work} & \textbf{event}& \textbf{time} & \textbf{metric}\\
        \Xhline{1pt}
\rule{0pt}{2.6ex} 
creative work&event name&date or period&cardinal number\\
creative work -> album&event name -> election&operating hours&measurement quantity\\
creative work -> book&event name -> geographical phenomenon&sub-day time expression&monetary amount (with unit)\\
creative work -> magazine&event name -> others&well-defined time interval&percentage (with \%)\\
creative work -> media contents&event or activity name&year or time period&price description\\
creative work -> poem&&&process parameter\\
creative work -> song&&&ordinal number\\
&&&\\
        \Xhline{2pt}
    \end{tabular} }
    \caption{Full NER taxonomy of English entities in \texttt{B\textsuperscript{2}NERD}.}
    \label{tbl:en-taxonomy}
\end{table*}

\end{CJK}

\begin{CJK}{UTF8}{gbsn}
\begin{table*}
    \centering
    \resizebox{\textwidth}{!}{
    \begin{tabular}{c|c|c|c|c}
        \Xhline{2pt}
         \rule{0pt}{2.4ex} 
        \textbf{人物相关} & \textbf{地名相关} & \textbf{生物医学} & \textbf{物品相关} & \textbf{组织机构相关}\\

        \Xhline{1pt}
        \rule{0pt}{2.6ex} 
人名&乡镇&中医证候&产品名&组织机构名\\人名->体育经理&产品产地&中药功效&产品名->乐器&组织机构名(不带地理政治实体)\\人名->其它&兴趣点&医学检查项目&产品名->交通工具&组织机构名->体育团队\\人名->政治人物&区县&医学检查项目->实验室检验项目&产品名->其它&组织机构名->公共公司\\人名->歌手&单元号&医学检查项目->影像检查&产品名->待售产品&组织机构名->公司\\人名->犯罪嫌疑人&国家&医学检查项目的名称或泛称&产品名->服装&组织机构名->其它\\人名->神职人员&地名&医学检测结果&产品名->相关产品&组织机构名->政府机构\\人名->科学家&地名->其它&医疗设备&产品名->金融产品&组织机构名->汽车制造商\\人名->艺术家&地名->景点&医院科室&产品名->食品&组织机构名->私营公司\\人名->被害人&地名->目的地&微生物名&产品名->食品或饮品&组织机构名->航空航天制造商\\人名->运动员&地名->设施&检查或治疗程序&产品名->饮品&组织机构名->银行\\人名或昵称&地名->起点&毒品或毒品成分名&产品型号&组织机构名->音乐团体\\人物&地名->车站&治疗措施(不含手术)&产品系列名&组织机构泛称\\人物或团体名&地名完整描述&治疗措施(含药物)&产品配件&组织机构的名称或泛称\\人群泛称&地名或地理政治实体&治疗措施->手术&品牌名&\\人群类别&地点(不带地理政治实体)&治疗措施描述&商品名&\\国籍&地点泛称&生化成分&手机型号&\\地名->人类居住地&地点的名称或泛称&疾病名&涉案物品完整描述&\\头衔&地理政治实体&疾病类别&物品的名称或泛称&\\收件人或收件单位&地理政治实体泛称&疾病诊断&食品类别&\\民族&城市&病因&&\\用户群体&子兴趣点&症状&&\\籍贯&开发区&症状或体征&&\\职业或职位&房间号&症状或体征描述&&\\联系人名&普通辅助定位词&细胞类型&&\\&村民小组&给药方式&&\\&村社&药品名&&\\&楼层号&药物&&\\&楼栋号&药物剂型&&\\&次级道路&药物名&&\\&次级道路门牌号&药物性味&&\\&省份&药物成分&&\\&自定义目的地&药物的名称或类别&&\\&距离辅助定位词&药物类别&&\\&路口&解剖学实体(非标准)或细胞名&&\\&道路&解剖部位&&\\&门牌号&解剖部位(含动植物)&&\\&&身体部位&&\\&&身体部位或身体物质&&\\
&&&&\\
\Xhline{2pt}
        \rule{0pt}{2.4ex} 
        \textbf{作品相关} & \textbf{度量相关}& \textbf{教育相关} & \textbf{时间相关} & \textbf{其它}\\
        \Xhline{1pt}
\rule{0pt}{2.6ex} 
作品名&产品规格尺寸&专业名&产品使用期限&产品主题\\作品名->影像作品&剂量&教育相关实体&年龄段&产品使用场所\\作品名->文字作品&度量&教育背景&持续时间&产品功能描述\\作品名->游戏作品&数值&数学概念&时间完整表述&产品味道\\作品名->电视节目&频率&解题原理或方法&时间状语&产品图案\\作品名->艺术作品&程度&&时间相关信息&产品型号编号\\作品名->软件&重量&&时间相关表述&产品外形描述\\作品名->音乐作品&财产价值(带币种)&&&产品服务描述\\作案工具&销赃金额(带币种)&&&产品材料类型\\副作用&现金或转账金额(带币种)&&&产品款式\\歌曲语言&&&&产品气味\\音乐主题&&&&产品用途\\音乐列表类别&&&&产品订阅类型\\音乐风格&&&&产品配置参数\\&&&&其它\\&&&&形容词评价\\&&&&情感类型\\&&&&抽象概念\\&&&&电脑硬件规格\\&&&&电话号码\\&&&&证书文档\\&&&&金融术语\\&&&&音乐场景\\&&&&颜色\\
&&&&\\
        \Xhline{2pt}
    \end{tabular} }
    \caption{Full NER taxonomy of Chinese entities in \texttt{B\textsuperscript{2}NERD}.}
    \label{tbl:zh-taxonomy2}
\end{table*}

\end{CJK}

\end{document}